\documentclass{article}

\PassOptionsToPackage{square,sort,comma,numbers}{natbib}
\usepackage[final]{nips_2017}

\usepackage[utf8]{inputenc} 
\usepackage[T1]{fontenc}    
\usepackage{hyperref}       
\usepackage{url}            
\usepackage{booktabs}       
\usepackage{amsfonts}       
\usepackage{nicefrac}       
\usepackage{microtype}      

\usepackage{xcolor}
\usepackage{graphicx}
\usepackage{subcaption}
\usepackage{float}
\usepackage[justification=raggedright]{caption}	
\usepackage{lscape}                                         

\usepackage[lined,ruled,linesnumbered]{algorithm2e}

\usepackage{booktabs}                   
\usepackage{multirow}

\usepackage{paralist}
\usepackage{enumitem}

\usepackage{bm}                          
\usepackage{epsfig}                      
\usepackage{graphicx}                  
\usepackage{times}
\usepackage{mathptmx}
\usepackage{mathtools}
\usepackage{amssymb,amsmath}   

\usepackage{units}
\usepackage{color}

\usepackage{comment}

\usepackage{url}  
\usepackage{xspace}
\usepackage{setspace}

\def\etal{et~al.~}			 
\def\eg{e.g.,~}               
\def\ie{i.e.,~}               

\def\xnormal{$x^\mathrm{normal}_{t} $}
\def\xadv{$x^\mathrm{adv}_{t} $}
\def\xt{$x_{t} $}

\newlength\paramargin
\newlength\figmargin
\newlength\secmargin

\newlength\figwidth

\newcommand{\secref}[1]{Section~\ref{sec:#1}}
\newcommand{\figref}[1]{Figure~\ref{fig:#1}}

\long\def\ignorethis#1{}
\newcommand {\jiabin}[1]{{\color{blue}\textbf{Jia-Bin: }#1}\normalfont}

\def\tmp#1{\textcolor{red}{#1}}

\newcommand{\tb}[1]{\textbf{#1}}

\providecommand{\shortcite}[1]{\cite{#1}}

\newcommand{\ignore}[1]{}
\newcommand\crule[3][black]{\textcolor{#1}{\rule{#2}{#3}}}

\title{Detecting Adversarial Attacks on \\
Neural Network Policies with Visual Foresight}

\author{
  Yen-Chen Lin \\
  Virginia Tech \\
  \texttt{yclin@vt.edu} 
  \AND
  Ming-Yu Liu \\
  NVIDIA \\
  \texttt{mingyul@nvidia.com} 
  \And
  Min Sun \\
  National Tsing Hua University \\
  \texttt{sunmin@ee.nthu.edu.tw}
  \And
  Jia-Bin Huang \\
  Virginia Tech \\
  \texttt{jbhuang@vt.edu}
}

\begin{document}

\maketitle

\begin{abstract}
Deep reinforcement learning has shown promising results in learning control policies for complex
sequential decision-making tasks.
However, these neural network-based policies are known to be vulnerable to adversarial examples. 
This vulnerability poses a potentially serious threat to safety-critical systems such as autonomous vehicles.
In this paper, we propose a defense mechanism to defend reinforcement learning agents from adversarial attacks by leveraging an action-conditioned frame prediction module.
Our core idea is that the adversarial examples targeting at a neural network-based policy are not effective for the frame prediction model.
By comparing the action distribution produced by a policy from processing the current observed frame to the action distribution produced by the same policy from processing the predicted frame from the action-conditioned frame prediction module, we can detect the presence of adversarial examples. 
Beyond detecting the presences of adversarial examples, our method allows the agent to continue performing the task using the predicted frame when the agent is under attack. 
We evaluate the performance of our algorithm using five games in Atari 2600. 
Our results demonstrate that the proposed defense mechanism achieves favorable performance against baseline algorithms in detecting adversarial examples and in earning rewards when the agents are under attack.
%
\end{abstract}


\vspace{\secmargin}
\section{Introduction}\label{sec:intro}

Reinforcement learning algorithms that utilize Deep Neural Networks (DNNs) as function approximators have emerged as powerful tools for learning policies for solving a wide variety of sequential decision-making tasks, including playing games~\cite{mnih:human,silver:mastering} and executing complex locomotions~\cite{peng:terrain}.
These methods have also been extended to robotic research and shown promising results in several applications such as robotic manipulation, navigation, and autonomous driving~\cite{levine:end,zhu:target,bojarski:end}.

As inheriting the learning capability from DNNs, DNN-based policies also inherit the vulnerability to adversarial examples as shown in~\cite{sandy:adversarial,lin:tactics,behzadan:vulnerability}. 
Adversarial examples are inputs corrupted by small, imperceivable perturbations that are carefully crafted for producing dramatic changes in outputs of a machine learning system. 
Adversarial examples have been extensively studied in the context of image classification~\cite{szegedy:intriguing,goodfellow:explaining}, object detection and semantic segmentation~\cite{xie2017adversarial,cisse2017houdini}, and their existences in the physical world were recently revealed by Kurakin~\etal\shortcite{kurakin:adversarial}. 
For DNN-based policies, the vulnerability to adversarial attack~\cite{lin:tactics,sandy:adversarial,behzadan:vulnerability} can potentially pose a significant danger to safety-critical applications including self-driving cars and human-robot collaboration platforms. 
Hence, developing an effective defensive mechanism against adversarial examples is an important problem.

\begin{figure}[t!]
\centering
\includegraphics[trim=0.0in 0.0in 0.0in 0in,width=\linewidth]{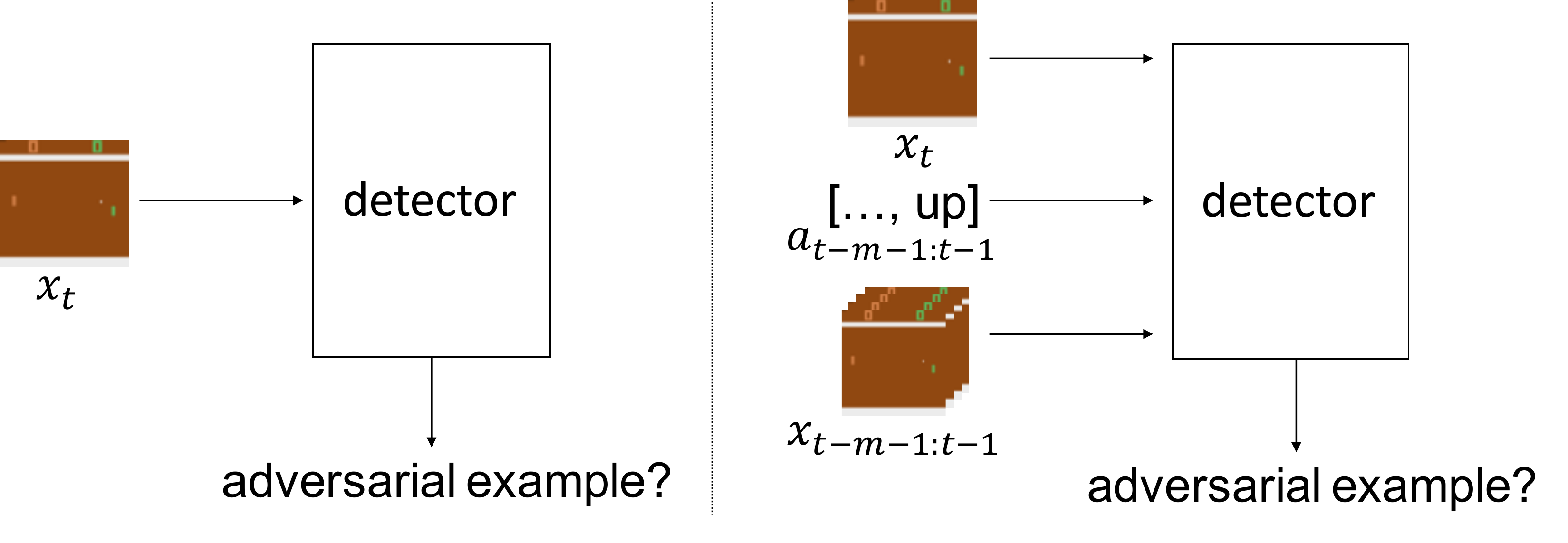}
\\
\begin{minipage}{0.49\linewidth} \centering
(a) Single input
\end{minipage} \hfill
\begin{minipage}{0.49\linewidth} \centering
(b) Multiple inputs (ours)
\end{minipage}
\caption{
\tb{Inputs of an adversarial examples detector.} Let $x_{t}$ be current input frame, $x_{t-m-1:t-1}$ be previous $m$ input frames, and $a_{t-m-1:t-1}$ be previous $m$ executed actions.
(a) Existing defense approaches for detecting adversarial examples are mainly developed in the context of image classification. As a result, these defense approaches take only one single frame as input per defense act.
(b) In contrast, the proposed defense leverages the temporal coherence of multiple inputs and the executed actions over time to detect adversarial attacks.
%
}
\label{fig:comparison}
\end{figure}

In this paper, we propose a defense mechanism specially designed to detect adversarial attacks on DNN-based policies and provide action suggestions for agents when under attacks. 
As shown in~\figref{comparison}, unlike existing defensive methods developed for image classification that only use information in \emph{one image} per defense act, the proposed method leverages temporal coherence of \emph{multiple frames} in sequential decision-making tasks.
Specifically, we train a visual foresight module---an action-conditioned frame prediction model that predicts the current frame based on the past observed frames and actions.
In general, future frame prediction is a difficult problem. 
However, in domains where the system dynamics are known or can be accurately modeled (e.g., robotic manipulation~\cite{finn:deep} or video games~\cite{oh:action}), we can train an accurately visual foresight module.
With the visual foresight module, we can detect adversarial examples by comparing 1) the action distribution generated by the policy using the current observed frame and 2) the action distribution generated by the same policy but using the predicted frame. 
When adversarial perturbations are present in the current frame, the two action distributions will be very different, and their similarity provides a measure of the presence of adversarial attacks.
Once adversarial attacks are detected, an agent could thus act based on the predicted frame instead of the observed frame. 
Of course, adversarial examples could also be present in the previous frames, which are used by the visual foresight module. 
However, since the adversarial examples are crafted to fool the policy, not the visual foresight module, its impact to the visual foresight module is small. 
Consequently, the visual foresight module can still predict plausible current frame conditioning on previous (and potentially attacked) frames and the executed actions.

The proposed defense has the following three main merits. 
First, our approach produces action suggestions using the visual foresight model for the agent under attack to retain its performance. 
In contrast to existing defensive mechanisms for image classification where the goal is to \emph{reject} adversarial examples, our approach provides ways for the agent to \emph{maneuver} under adversarial attacks.
For example, with the proposed defensive method, a robot arm can safely manipulate an adversarially perturbed object instead of accidentally damaging it.
Second, our approach does not require adversarial examples to construct the defense. 
This suggests that our method is not specific to a particular adversarial attack as several existing methods~\cite{goodfellow:explaining,metzen:detecting} are.
Third, our method is model-agnostic and thus can potentially be used to protect a wide range of different DNN-based policies.

We validate the effectiveness of the proposed defensive mechanism in Atari game environments.
We show that our defense can reliably detect adversarial examples under consecutive attacks. 
Using the proposed action suggestions, our results show that the agent can retain the performance even when a large portion of time steps was attacked. 
Our approach demonstrates favorable results over several strong baseline detectors developed for detecting adversarial examples for image classification.
We also characterize the performance over the accuracy of the frame prediction model.


\tb{Our contributions:} 
\begin{itemize}
\item To the best of our knowledge, we are the first to develop defensive methods against adversarial examples for DNN-based policies by leveraging temporal coherency.
\item The proposed defense method can not only detect adversarial examples but also provide action suggestions to retain agents' performance. 
On five games of Atari 2600, our defense mechanism significantly increases agents' robustness to reasonable limited-time attack.
\item Our method is agnostic to both the model of the policy and the adversary. 
Therefore, our approach can be applied to protect a wide variety of policies without collecting adversarial examples ahead of time.
\item We discuss several practical aspects of adversarial attacks on DNN-based policies and limitations of our approach.
\end{itemize}

\begin{figure*}[t!]
\centering
\includegraphics[trim=0.0in 0.0in 0.0in 0in,width=\textwidth]{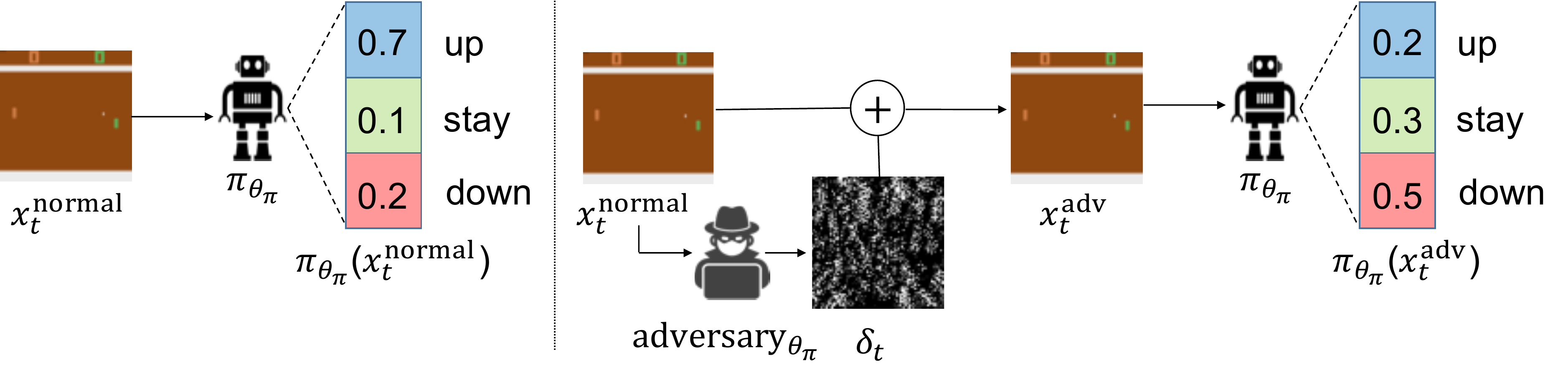}
\\
\hfill
\begin{minipage}{0.35\textwidth} \centering
(a) \emph{without} adversary
\end{minipage} 
\hfill
\begin{minipage}{0.64\textwidth} \centering
(b) \emph{with} adversary
\end{minipage}
\hfill
\caption{
\tb{Threat model}. 
(a) A well-trained agent can process the normal inputs and choose appropriate actions to achieve high cumulated rewards. 
(b) In our threat model, the adversary is not aware of the defense, but is aware of the parameters $\theta_{\pi}$ of the policy. The adversary can thus apply malicious crafted perturbation $\delta_{t}$ to degrade the run-time performance of a trained agent $\pi_{\theta_{\pi}}$. In this example, the adversary fools the agent to produce an incorrect action distribution $\pi_{\theta_{\pi}}(x^{\mathrm{adv}}_{t})$.
}
\label{fig:threat_model}
\end{figure*}

\vspace{\secmargin}
\section{Related Work}\label{sec:related}

\paragraph{Adversarial examples.} 
Szegedy \etal \shortcite{szegedy:intriguing} first showed an intriguing property that deep neural networks are vulnerable to adversarial examples---maliciously designed inputs to cause a model to make a mistake in the context of image classification tasks. 
Several recent efforts have been devoted to developing more sophisticated attacks~\cite{goodfellow:explaining,moosavi:deep,papernot:limitations,carlini-wagner:towards,kurakin:adversarial} and demonstrating the feasibility of attacking real-world machine learning systems using adversarial examples~\cite{papernot:practical,liu:delving}.
Beyond classification tasks, it has also been shown that other applications of deep learning such as deep neural network policies and deep generative models are also vulnerable to adversarial examples~\cite{lin:tactics,sandy:adversarial,kos:delving,kos:adversarial}. 
As adversarial examples pose potential security threats to machine learning based applications, designing defensive mechanisms that can protect against such adversaries is of great importance.
Our goal in this paper is to develop a defense to protect deep neural network policies.

\vspace{\paramargin}
\paragraph{Defenses against adversarial examples.} 
Existing defense mechanisms mostly focus on image classification tasks.
These approaches can be grouped into two main categories. 
The first category proposes enhanced training procedures to make a trained model robust to adversarial examples. 
Methods in this category include adversarial training~\cite{goodfellow:explaining,kurakin:adversarial} and defensive distillation~\cite{papernot:distillation,papernot:extending}. 
Recently, it has been shown that this category of methods is not effective against optimization-based adversarial attacks~\cite{carlini-wagner:towards}. 
The second category focuses on detecting and rejecting adversarial perturbations
~\cite{metzen:detecting,xu:feature,liang:detecting,Bhagoji:dimensionality,feinman:detecting,gong:adversarial,grosse:statistical,li:adversarial}. 
While these methods are promising in detecting adversarial examples, they do not provide a mechanism to carry out the recognition task under adversarial attacks.

Our method differs from existing work in two main aspects. 
First, existing defense approaches focus on image classification tasks, and thus the mechanism is based on a \emph{single} image. 
In contrast, we focus on the problem of detecting adversarial examples in sequential decision-making tasks by exploiting the temporal coherence of \emph{multiple} input observations. 
Second, in addition to detecting adversarial examples, our method also provides action suggestions for the agents under attack and help the agents retain its performance. 
Such a property is particularly important in sequential decision-making applications.

\vspace{\paramargin}
\paragraph{Video prediction.} Predicting future frames given the current observed frames~\cite{srivastava2015unsupervised,oh:action,mathieu:deep,finn:unsupervised,finn:deep,xue2016probabilistic,villegas2017decomposing} itself is an important problem. 
Among the future frame prediction works, the works on action-conditioned next frame prediction methods~\cite{oh:action,finn:unsupervised,finn:deep} are particularly suitable for predicting the future frame that could be observed by an agent. 
Our work applies an action-conditioned frame prediction model for detecting adversarial examples in sequential decision tasks.

\vspace{\secmargin}
\section{Preliminaries}\label{sec:preliminaries}


\vspace{\paramargin}
\paragraph{Reinforcement learning.}
Reinforcement learning concerns learning a policy for an agent to take actions to interact with an environment in order to maximize the cumulative reward. 
Here, the environment is modeled as a Markov Decision Process (MDP). 
An MDP $M$ is a 5-tuple $(S, A, P, R, \gamma)$, where $S$ is a set of states, $A$ is a set of actions, $P(s_{t+1} \mid s_{t}, a_{t})$ is the transition probability from state at the $t$-th step $s_{t} \in S$ to $s_{t+1} \in S$ after performing the action $a_{t} \in A$, $R(r_{t+1} \mid s_{t}, a_{t})$ is the probability of receiving reward $r_{t+1}$ after performing action $a_{t}$ at state $s_{t}$,  and $\gamma \in [0, 1]$ is a future reward discounted rate. 
The goal of the agent is to learn a policy $\pi : S \rightarrow A$, which maps a state to an action, to maximize the expected cumulative reward collected in an episode of $T$ steps $s_{0}, a_{0}, r_{1}, s_{1}, a_{1}, r_{2}, ..., s_{T}$. 
When a complete state $s_{t}$ in an environment is not visible, the agent take as input the observation $x_{t}$ of the environment.
\vspace{\paramargin}
\paragraph{Threat model.}
\figref{threat_model} illustrates our threat model. 
At each time step $t$, an adversary can use adversarial examples generation algorithms to craft malicious perturbation $\delta_{t}$ and apply it to the observation $x_{t}$ to  create perturbed observation $x^{adv}_{t}$. Taking $x^{adv}_{t}$ as input, the agent produces an incorrect action distribution $\pi_{\theta_{\pi}}(x^{\mathrm{adv}}_{t})$ which may degrade its performance.
Specifically, we consider the white-box attack where the adversary has access to the parameters of the policy network to craft perturbed observation. 
We further assume that the adversary is static (\ie the adversary is unaware of the defense mechanism being used).

\vspace{\secmargin}
\section{Adversarial Example Detection}\label{sec:method}

Our goal is to detect the presence of adversarial attack at each time step in a sequential decision task. 
We denote the observation perceived by the agent at time step $t$ as \xt. 
In our threat model (\figref{threat_model}), the observation \xt can either be a normal (clean) observation \xnormal~or a maliciously perturbed observation \xadv.
Compared to existing defense methods developed for image classification tasks that take a single image as input, our method leverages the past $m$ observations $x_{t-m:t-1}$ and actions $a_{t-m:t-1}$, as illustrated \figref{comparison}.
In addition to detecting adversarial attacks, we show that our method can help the agent maneuver under adversarial attacks. 
Below we describe our approach in detail.

\begin{figure*}[t!]
\centering
\includegraphics[trim=0.0in 0.0in 0.0in 0in,width=\textwidth]{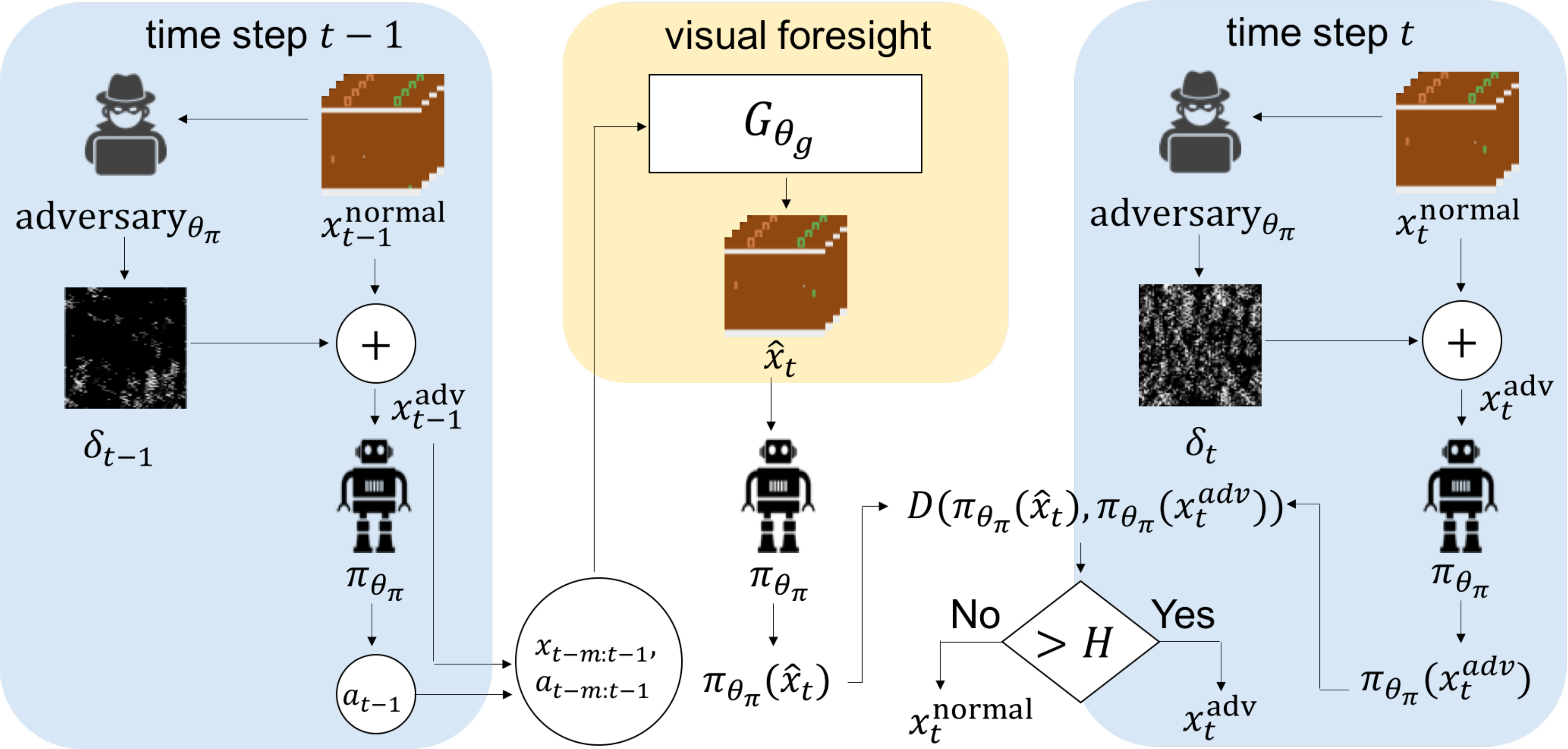}
\vspace{\figmargin}
\caption{
\tb{Algorithm overview.} 
The illustration here describes the scenarios where an adversary applies a consecutive attacks on the agent $\pi_{\theta_\pi}$. At the time step $t-1$ and $t$, the agent perceives the maliciously perturbed inputs $x^{\mathrm{adv}}_{t-1}$ and $x^{\mathrm{adv}}_{t}$ and may produces action distributions that lead to poor performance. With the incorporation of the visual foresight module that predicts the current frame $\hat{x_t}$ given the previous observations and executed actions, we can produce the action distribution $\pi_{\theta_{\pi}}(\hat{x_{t}})$ based on the predicted observations. By computing the similarity between the two action distributions $\pi_{\theta_{\pi}}(x_{t})$ and $\pi_{\theta_{\pi}}(\hat{x_{t}})$, we can determine the presence of the adversarial example if the distance $D(\pi_{\theta_{\pi}}(\hat{x_{t}}), \pi_{\theta_{\pi}}(x_{t}))$ exceeds a pre-defined threshold $H$. \ignore{Note that here we simplify the inx6puts to the agent to avoid clutter (\eg the agent takes the observations and executed actions in multiple time steps as inputs).} 
%
}
\label{fig:overview}
\end{figure*}

\vspace{\secmargin}
\subsection{Detecting Adversarial Attacks}
\label{sec:method_detection}

\figref{overview} illustrates the pipeline of the proposed method. 
At time step $t$, the action-conditioned frame prediction model $G_{\theta_g}$ takes $m$ previous observations $x_{t-m:t-1}$ and corresponding $m$ actions $a_{t-m:t-1}$ as input to predict the current observation $\hat{x}_t$. 
Given a normal observation \xnormal~at the current time step $t$, the action distribution that the agent uses to sample an action from is $\pi_{\theta_{\pi}}(x^\mathrm{normal}_{t})$, which should be similar to the action distribution of $\pi_{\theta_{\pi}}(\hat{x_{t}})$ from the predicted frame. 
On the other hand, if the current input is adversarially perturbed, that is the agent observed \xadv~instead of \xnormal, then the resulting action distribution $\pi_{\theta_{\pi}}(x^\mathrm{adv}_{t})$ should differ a lot from $\pi_{\theta_{\pi}}(\hat{x_{t}})$ because the goal of the adversary is to perturb the input observation \xt~to cause the agent to take a different action. 
Therefore, we can use the similarity between the two action distributions to detect the presence of adversarial attacks.
Specifically, we compute $D(\pi_{\theta_{\pi}}(\hat{x_{t}}), \pi_{\theta_{\pi}}(x_{t}))$, where $D(\cdot, \cdot)$ is a distance metric measuring the similarity. 
In this work, we use the $\ell_1$ distance as the distance metric but note that other alternatives such as chi-squared distance or histogram intersection distance can be used as well. 
We label \xt~as an adversarial example when $D(\pi_{\theta_{\pi}}(\hat{x_{t}}), \pi_{\theta_{\pi}}(x_{t}))$ exceeds a predefined threshold $H$.

\vspace{\secmargin}
\subsection{Providing Action Suggestions} \label{sec:action_suggestions}

In addition to detecting adversarial attacks, our approach can provide action suggestions using the predicted frame from the visual foresight module. 
Specifically, when an attack is detected at time step $t$, we replace the action distribution $\pi_{\theta_{\pi}}(x^\mathrm{adv}_{t})$ (produced by processing the original, potentially attacked observations) to the action distribution $\pi_{\theta_{\pi}}(\hat{x_{t}})$.

\vspace{\secmargin}
\subsection{Design of Frame Prediction Module} \label{method:visual_foresight}

Our visual foresight module is a frame prediction model that predicts the next observation \xt given the $m$ previous observations $x_{t-m:t-1}$ and corresponding $m$ actions $a_{t-m:t-1}$. 
We adopt the network architecture design in Oh \etal~\shortcite{oh:action}, which consists of three parts: 1) an encoder, 2) a multiplicative action-conditional transformation, and 3) a decoder.

\vspace{\secmargin}
\section{Experimental Results}\label{sec.Exp}
In this section, we validate the effectiveness of the proposed defense mechanism against adversarial examples.
After describing the implementation details in \secref{setup}, we will characterize the performance of attack detection in terms of precision and recall curves \secref{detection} and the effect of action suggestion using cumulative rewards under various levels of attack in \secref{reward}.
We will then discuss the effects on the performance over the accuracy of the frame prediction model in \secref{effect_frame_prediction}. 

\vspace{\secmargin}
\subsection{Experimental Setup} \label{sec:setup}

\paragraph{Datasets.} Following previous works~\cite{sandy:adversarial,lin:tactics} which demonstrate the vulnerability of DNN-based policy, we use five Atari 2600 games, \textsc{Pong}, \textsc{Seaquest}, \textsc{Freeway} \textsc{ChopperCommand} and \textsc{MsPacman} in our experiments to cover a wide variety of environments.
We choose the game environment to validate the effectiveness of the proposed defense mechanism because we can easily apply adversarial attacks on the inputs using existing adversarial example generation algorithms. 
Extension to a physical setting (\eg evaluating on robot manipulation or planning tasks) is non-trivial as it remains difficult to apply adversarial attacks at each time step in the physical world. 
We leave the extension to future work.

\vspace{\paramargin}
\paragraph{Training DNN-based policy.} We train our agent for each game using the DQN algorithm~\cite{mnih:human}.
We follow the input pre-processing steps and the neural network architecture as described in~\cite{mnih:human}.
Specifically, the input to the policy is a concatenation of the last $4$ frames, converted to grayscale and resized to $84 \times 84$.
For each game, we sample three random initializations of neural network parameters to train different policies. 
%
All of our experiments are performed using well-trained agents (\ie the policies achieve at least 80\% of the maximum reward for the last ten training iterations). 
While we use DQN~\cite{mnih:human} in our experiments, we note that our defense mechanism is agnostic to the policy used (\ie the same approach can be applied to alternative DNN-based policies such as  A3C~\cite{mnih:asynchronous} and TRPO ~\cite{schulman:trust}).

\vspace{\paramargin}
\paragraph{Training action-conditioned frame prediction model.} After training our agents, we use them to generate game-play video datasets using $\epsilon$-greedy policies which force the agents to choose a random action with 30\% probability.
%
We obtain 1,000,000 training frames for each game.
%
Following Mnih~\etal~\shortcite{mnih:human}, we choose actions once every 4 frames and thereby reduce the frame-rate of the video from 60fps to 15fps.
%
The number of available actions ranges from 3 to 18. 
We directly collect observations of agents and pre-process the images by subtracting the mean pixel values and dividing each pixel value by 255.
We adopt the CNN architecture and training procedure of~\cite{oh:action} to implement our action-conditioned frame prediction model.
Specifically, we adopt the curriculum learning scheme with three phases of increasing prediction step objectives of $1$, $3$ and $5$ steps, and learning rates of $10^{-4}$, $10^{-5}$ and $10^{-5}$.
We optimize the model using Adam optimizer~\cite{kingma:adam} with batch sizes of 32, 8, and 8. 
We conduct all of our experiments using OpenAI Gym~\cite{brockman:gym}. 
We will make the source code publicly available. 

\vspace{\secmargin}
\subsection{Detecting Adversarial Attacks} \label{sec:detection}

\paragraph{Adversarial example generation algorithms.} 
We evaluate our proposed method against three different adversarial example generation algorithms: FGSM~\cite{goodfellow:explaining}, BIM~\cite{kurakin:adversarial}, and Carlini~\etal~\shortcite{carlini-wagner:towards} with $\ell_\infty$-norm constraints $\epsilon= 0.003$.

\vspace{\paramargin}
\paragraph{Defense approaches.} 
We compare our method with three adversarial example detectors for image classification:
\begin{itemize}
\item Feature Squeezing~\cite{xu:feature}: We use the default parameters (\eg $2\times$2 median filters) as suggested by the authors. 
\item Autoencoder~\cite{dongyu:magnet}: We use the same architecture of our frame prediction model but remove the multiplicative action-conditional transformation.
We use the same amount of training data for action-conditioned frame prediction model to train the Autoencoder.
\item Dropout-based detection~\cite{feinman:detecting}: 
We train a DQN agent using dropout with probability $0.2$ before the fully connected layers.
We keep the dropout during testing and process each frame $30$ times to estimate the uncertainty of the policy prediction. 
We can then use this uncertainty measure to detect adversarial examples.
Our results show that the algorithm is not sensitive to the number of samples for estimating the uncertainty when it is greater than $20$. 
We thus use $30$ in our experiments.
%
\end{itemize}
%
%
%
%
Note that we do not use other detector-based defense mechanisms that require collecting adversarial examples for training their detectors.

\vspace{\paramargin}
\paragraph{Quantitative evaluation.} We present the quantitative evaluation of adversarial attacks detection. 
We characterize the level of the adversarial attack using ``attack ratio'', indicating the probability that an adversary launches an attack at each time step.

For each game, we run the agent five times where we set the attack ratio as 0.5 (\ie half of the time steps are attacked). 
We report the mean and the variance of the precision and recall curves in \figref{pr}. 
Note that we only count the ``successful'' adversarial examples that do change the action of the agent as positives.
In all five games under different types of attacks, our results show that the proposed adversarial detection algorithm compare favorably against other baseline approaches.

In \figref{action_distance}, we visualize the action distribution distance $D(\pi_{\theta_{\pi}}(\hat{x_{t}}), \pi_{\theta_{\pi}}(x_{t}))$ over the time steps in one trial for three games. 
Here, we apply consecutive attacks periodically for 200 frames. 
The visualization shows that we can leverage the distance to identify the adversarial attack even when the agent is consecutively attacked up to 200 frames.

\setlength{\figwidth}{0.19\textwidth}
\begin{figure*}[t!] 
\centering
\small

\rotatebox[origin=c]{90}{FGSM} \hfill
\begin{minipage}{\figwidth} \centering
\includegraphics[width=1.0\linewidth]{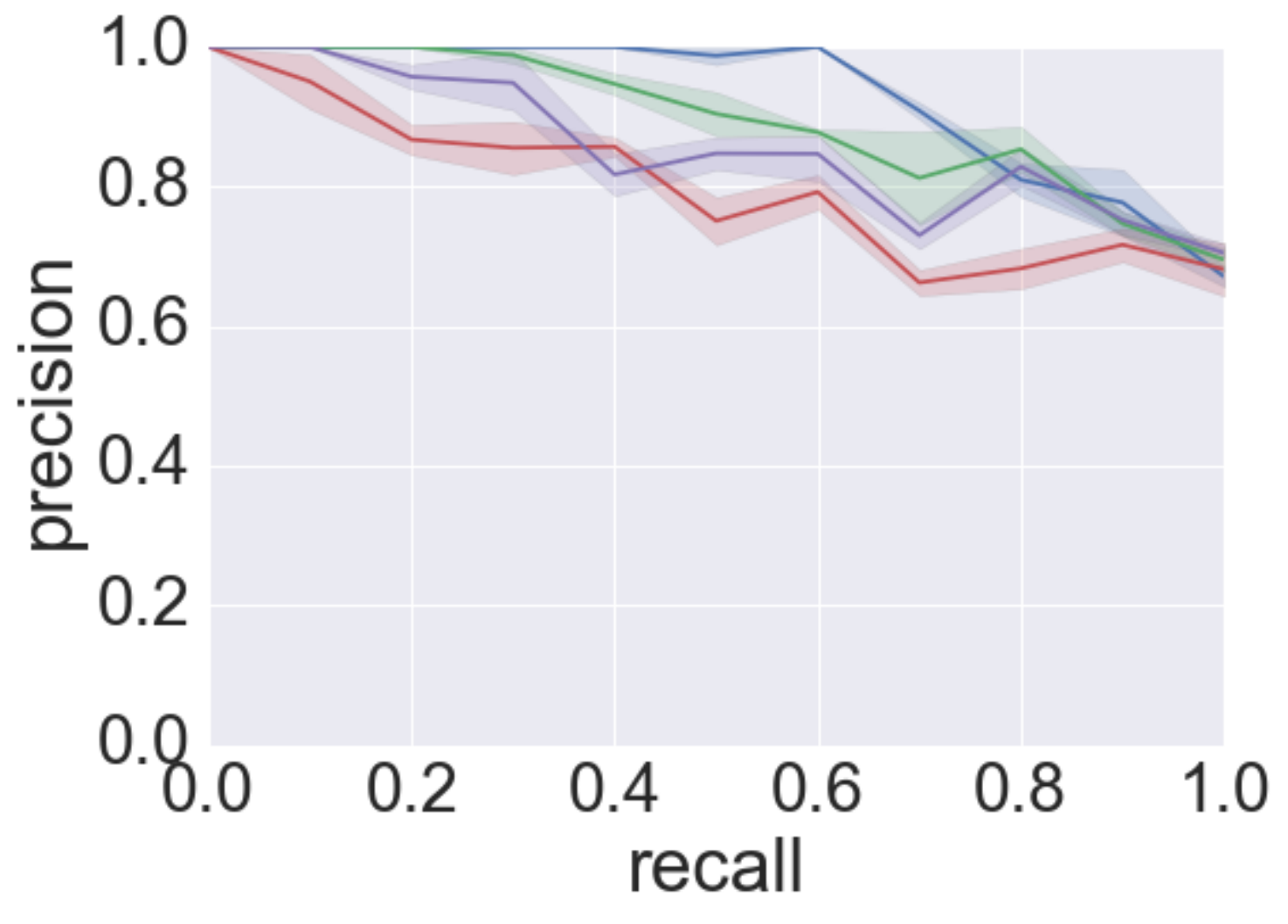}
\end{minipage}\hfill
\begin{minipage}{\figwidth} \centering
\includegraphics[width=1.0\linewidth]{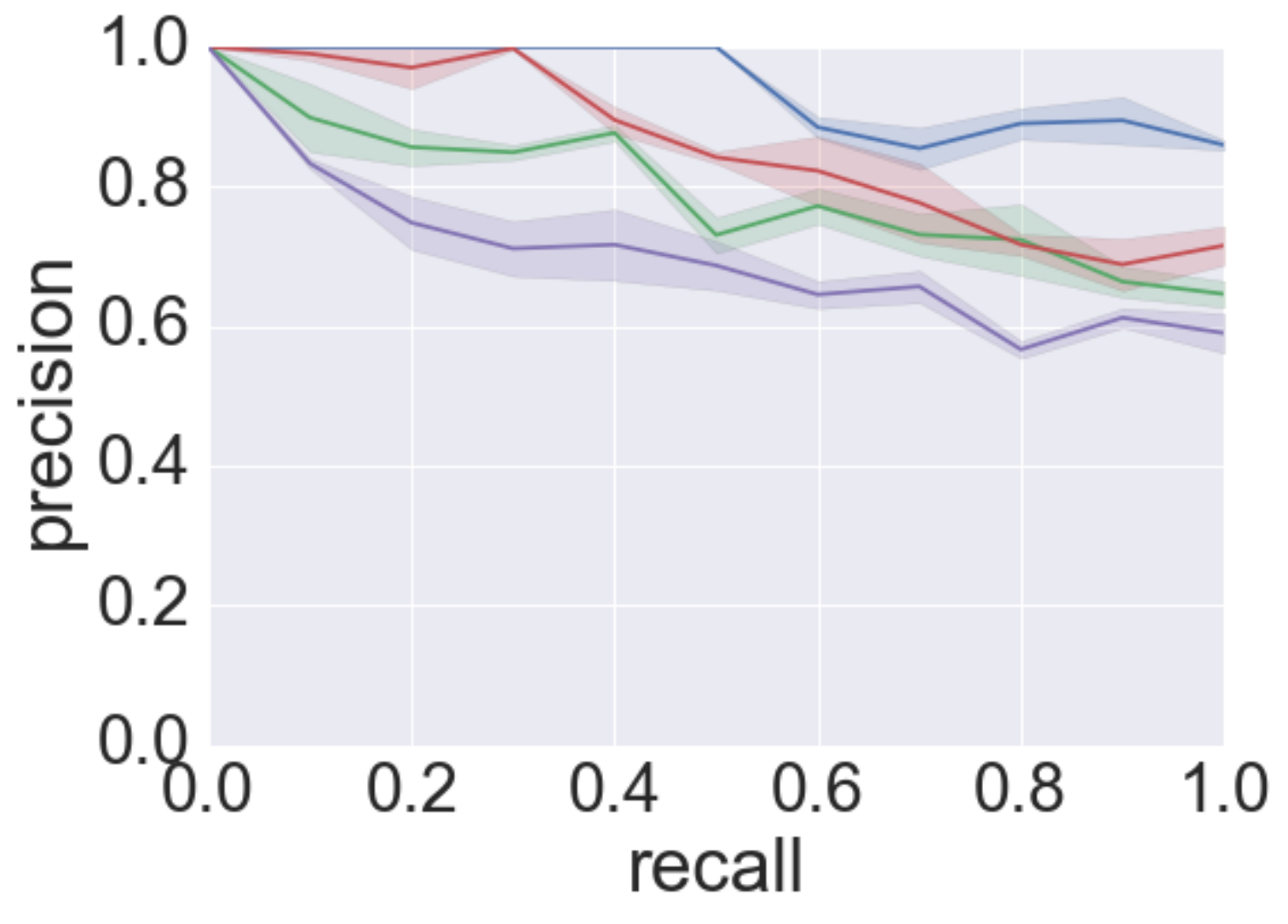} 
\end{minipage}\hfill
\begin{minipage}{\figwidth} \centering
\includegraphics[width=1.0\linewidth]{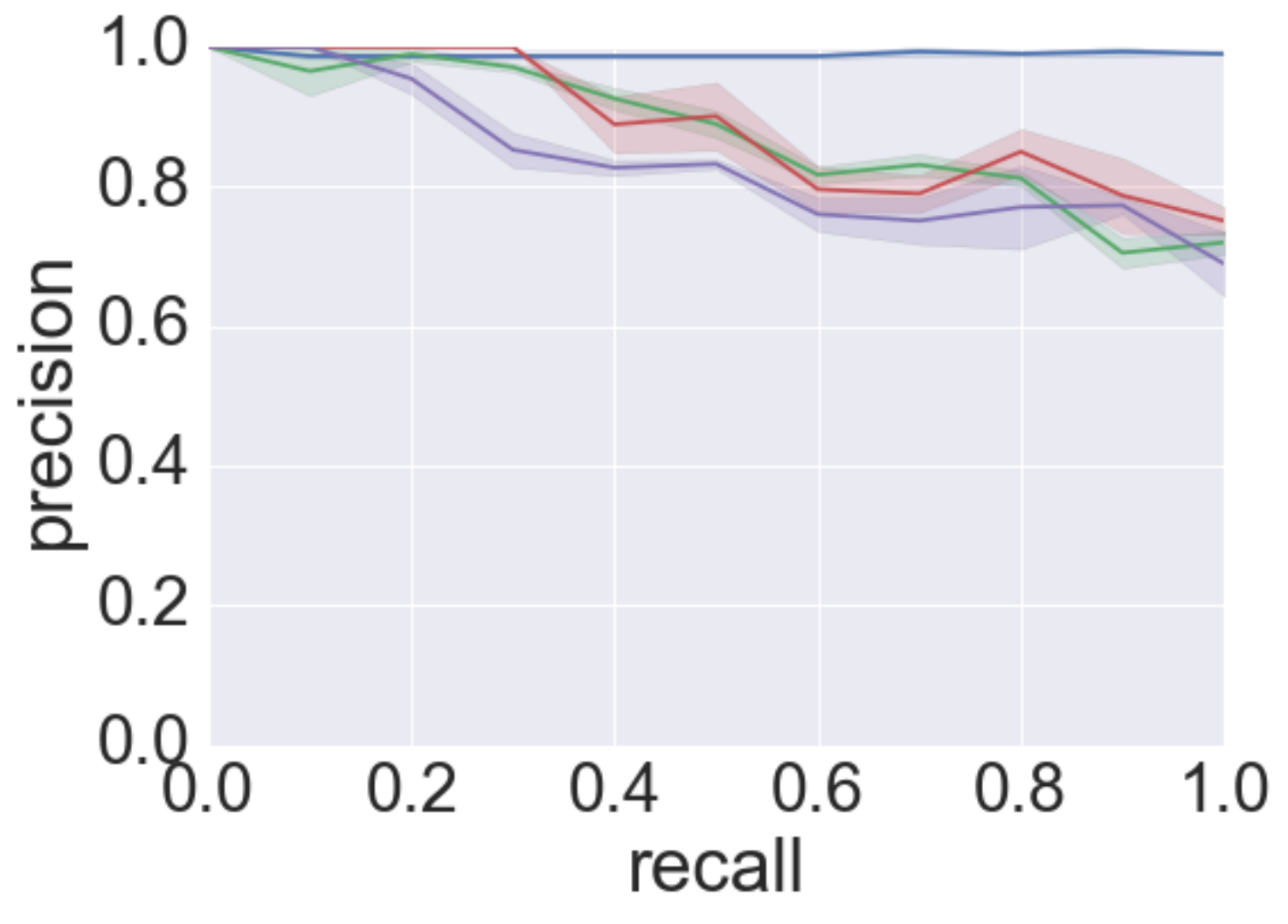} 
\end{minipage}\hfill
\begin{minipage}{\figwidth} \centering
\includegraphics[width=1.0\linewidth]{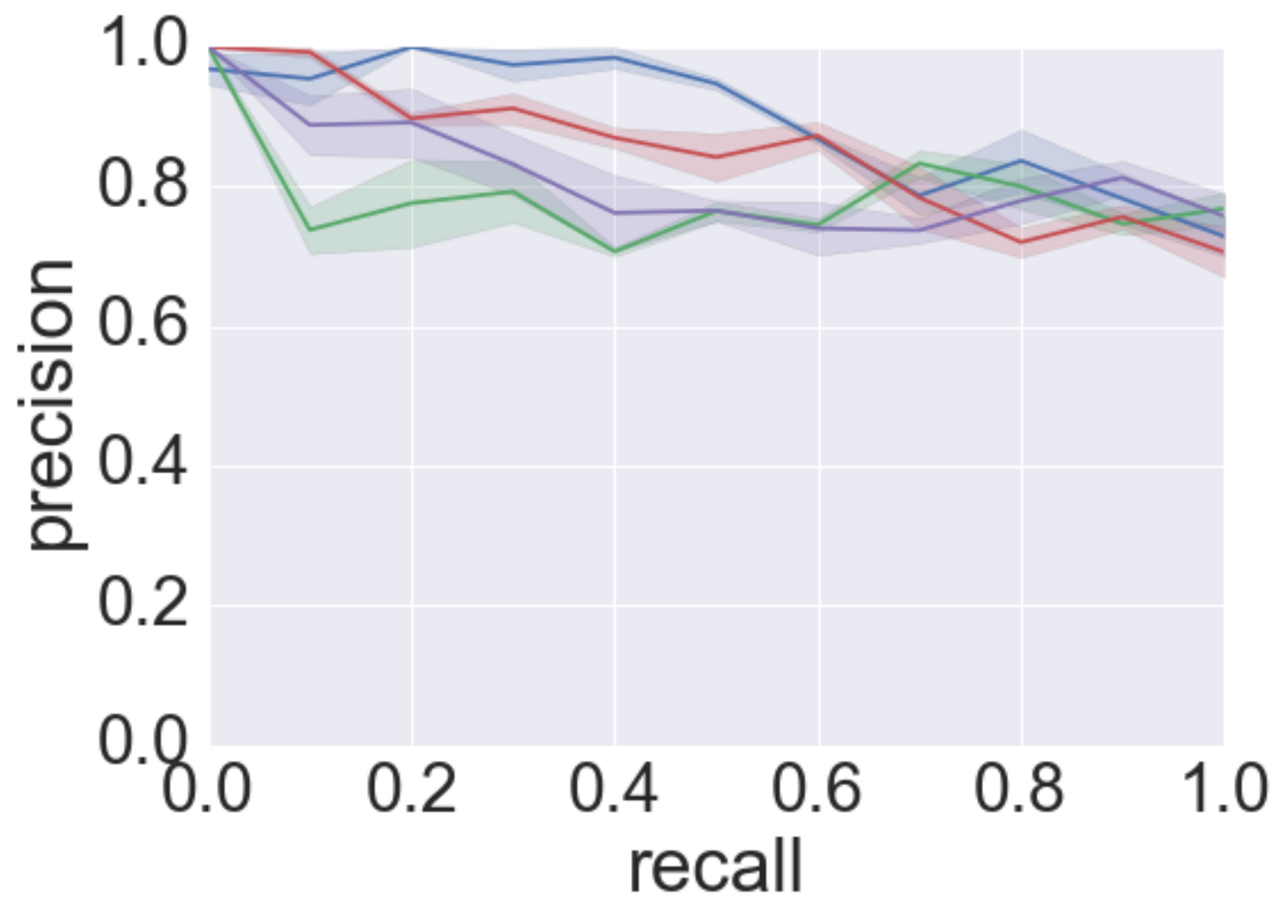} 
\end{minipage}\hfill
\begin{minipage}{\figwidth} \centering
\includegraphics[width=1.0\linewidth]{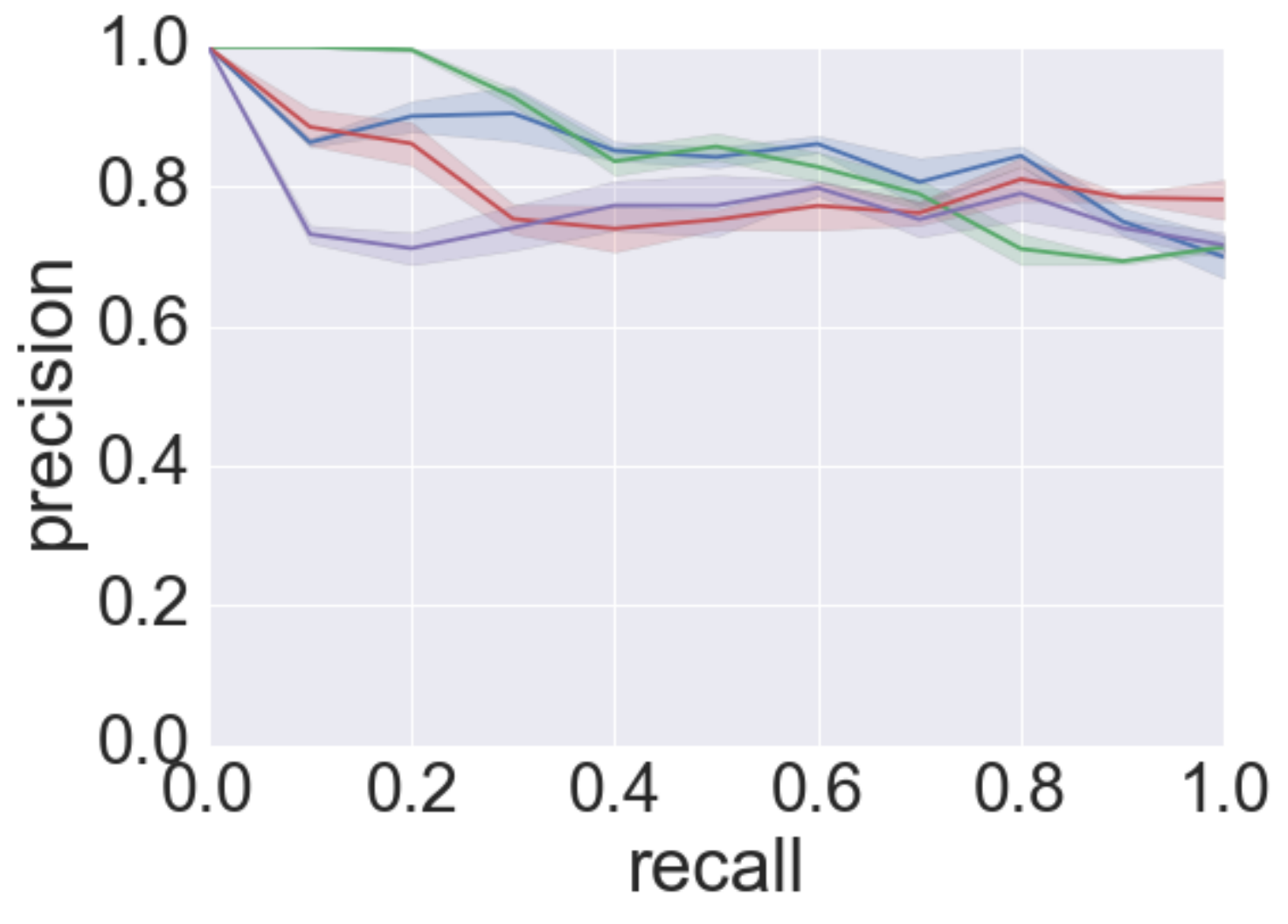}
\end{minipage}\hfill
\\
\rotatebox[origin=c]{90}{BIM} \hfill
\begin{minipage}{\figwidth} \centering
\includegraphics[width=1.0\linewidth]{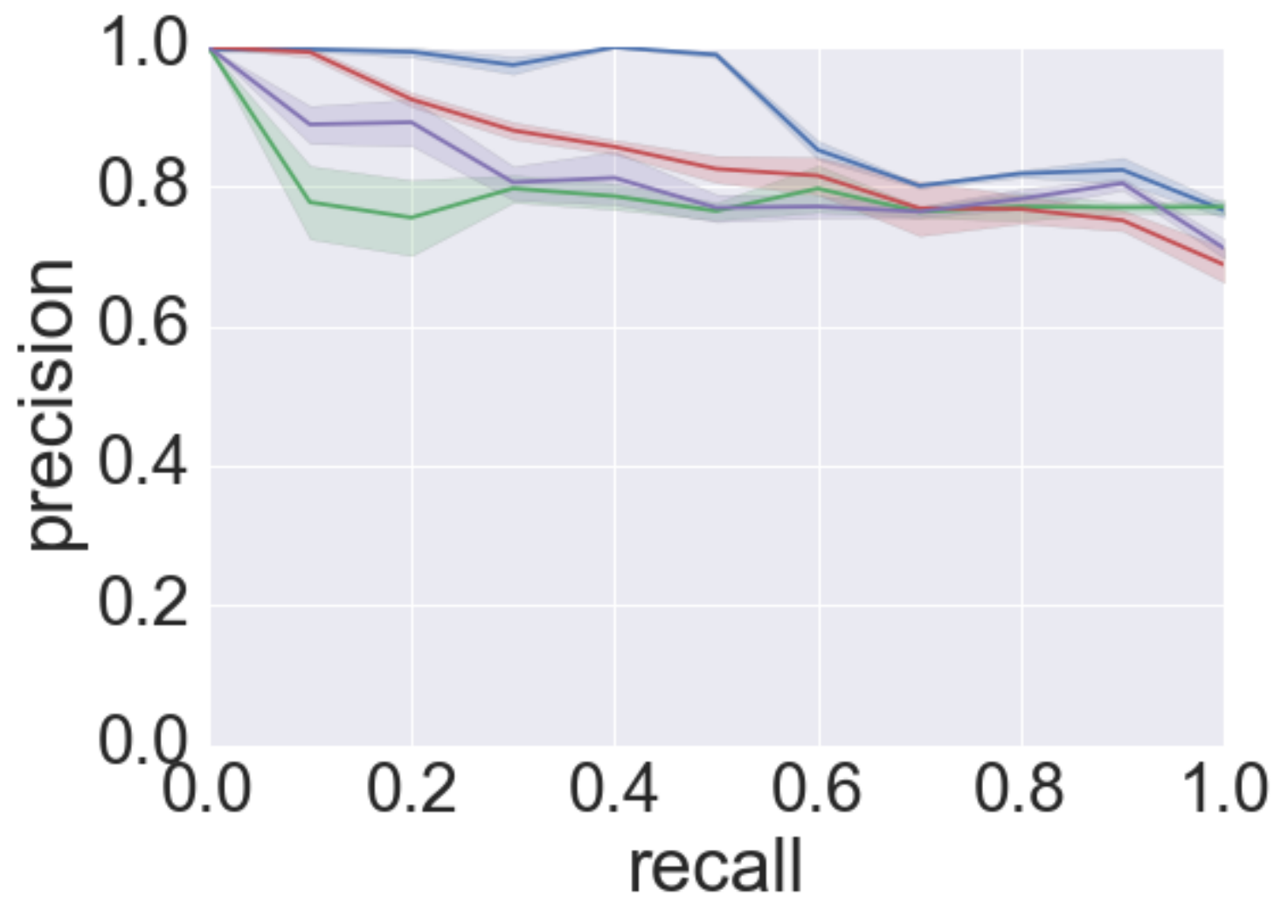}
\end{minipage}\hfill
\begin{minipage}{\figwidth} \centering
\includegraphics[width=1.0\linewidth]{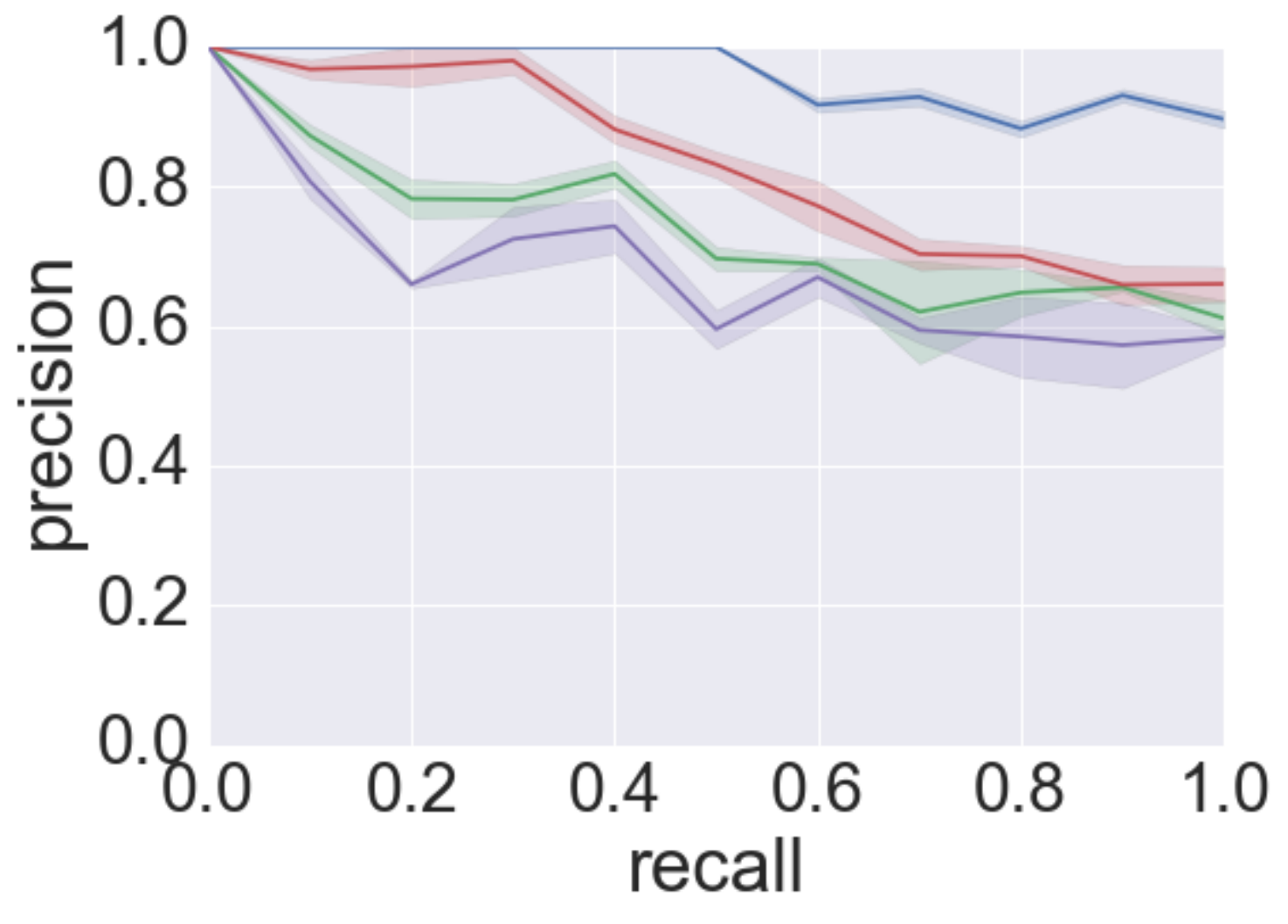} 
\end{minipage}\hfill
\begin{minipage}{\figwidth} \centering
\includegraphics[width=1.0\linewidth]{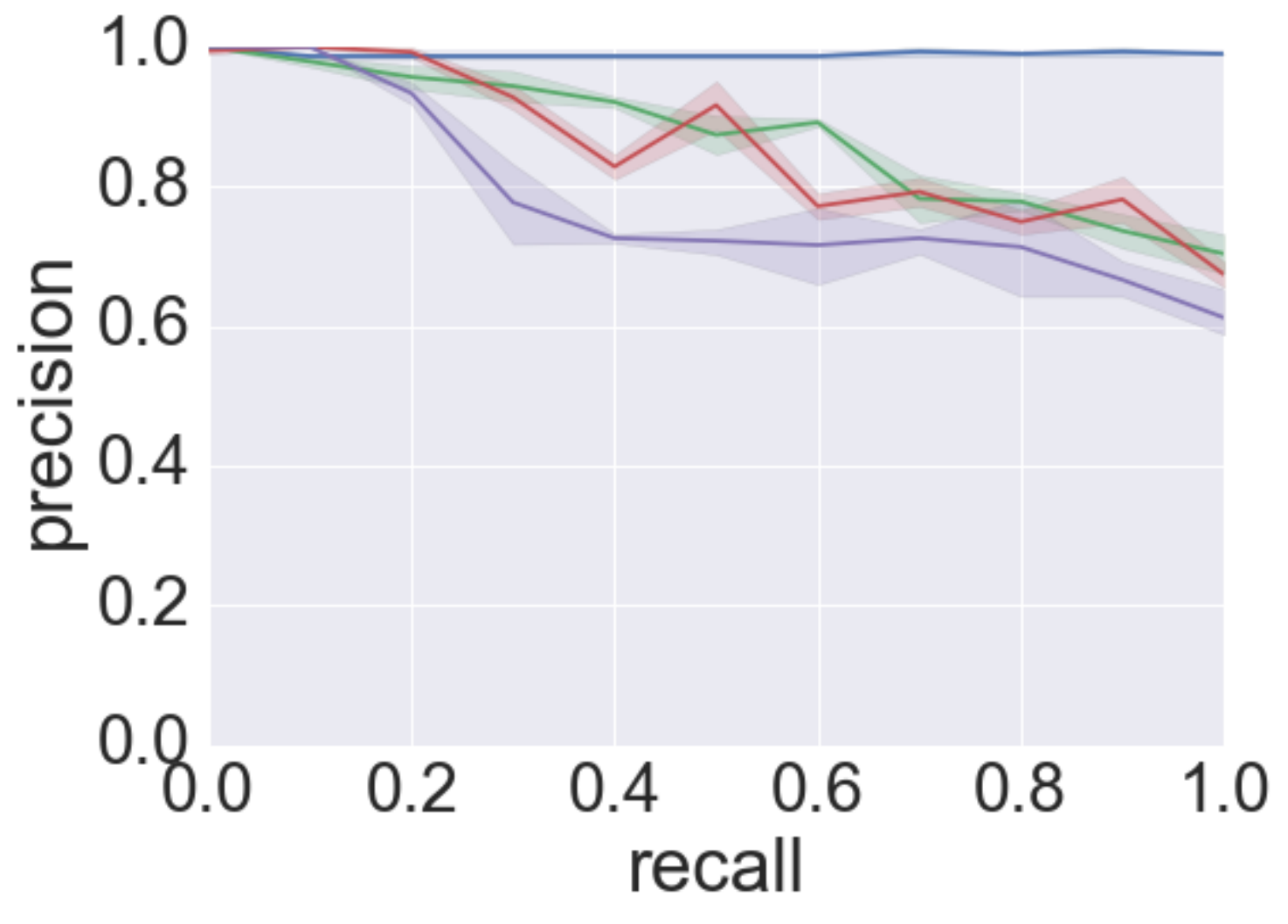} 
\end{minipage}\hfill
\begin{minipage}{\figwidth} \centering
\includegraphics[width=1.0\linewidth]{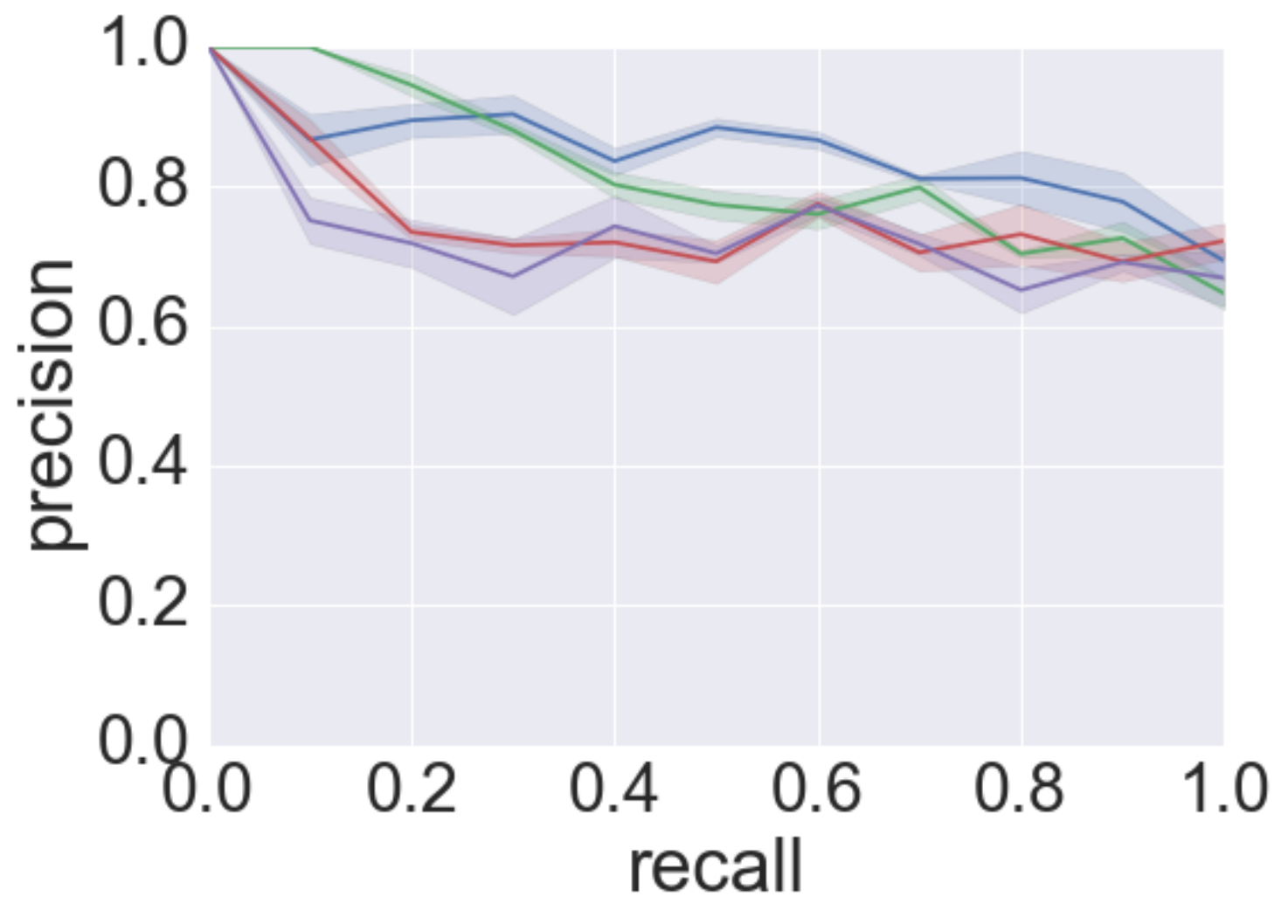} 
\end{minipage}\hfill
\begin{minipage}{\figwidth} \centering
\includegraphics[width=1.0\linewidth]{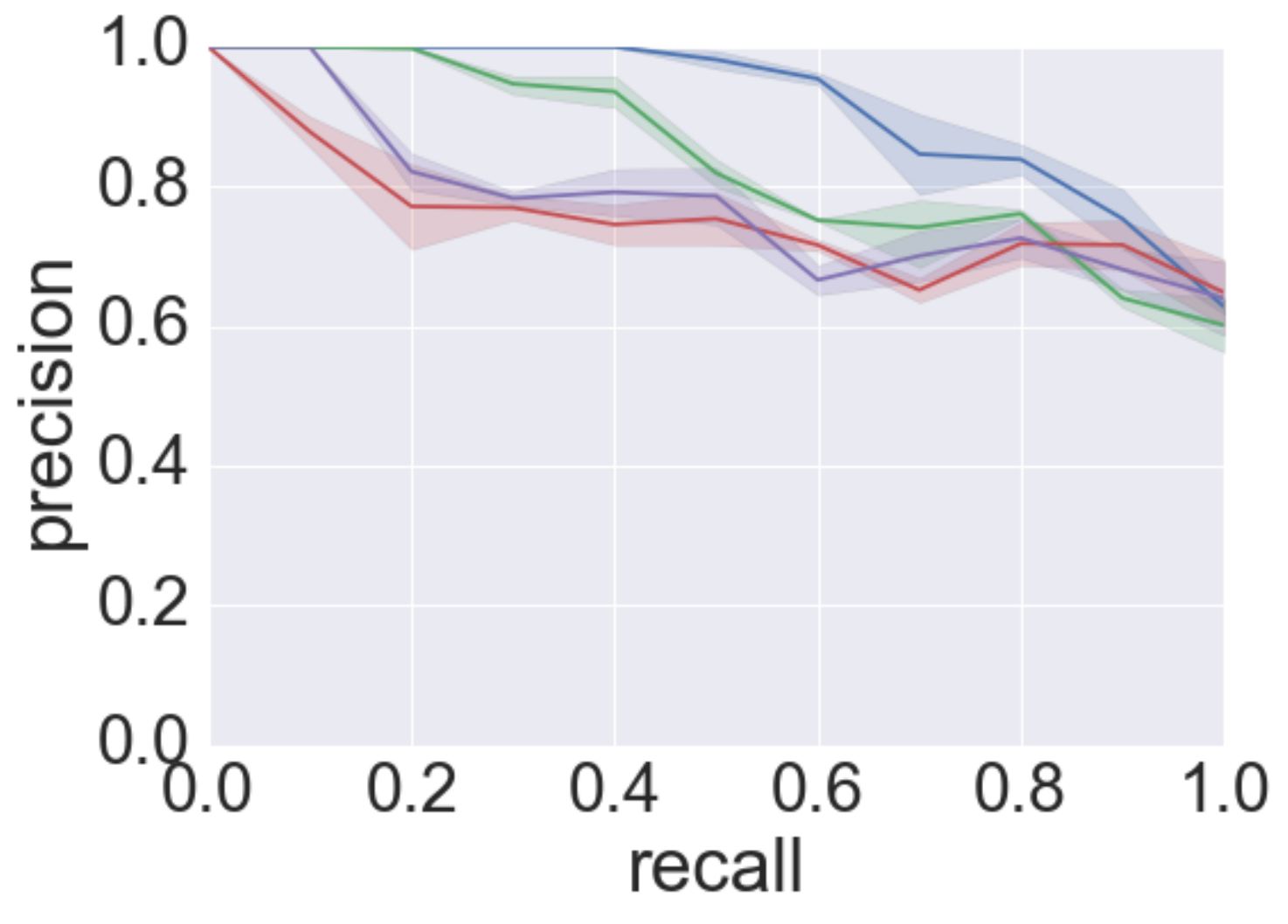}
\end{minipage}\hfill
\\
\rotatebox[origin=c]{90}{Carlini} \hfill
\begin{minipage}{\figwidth} \centering
\includegraphics[width=1.0\linewidth]{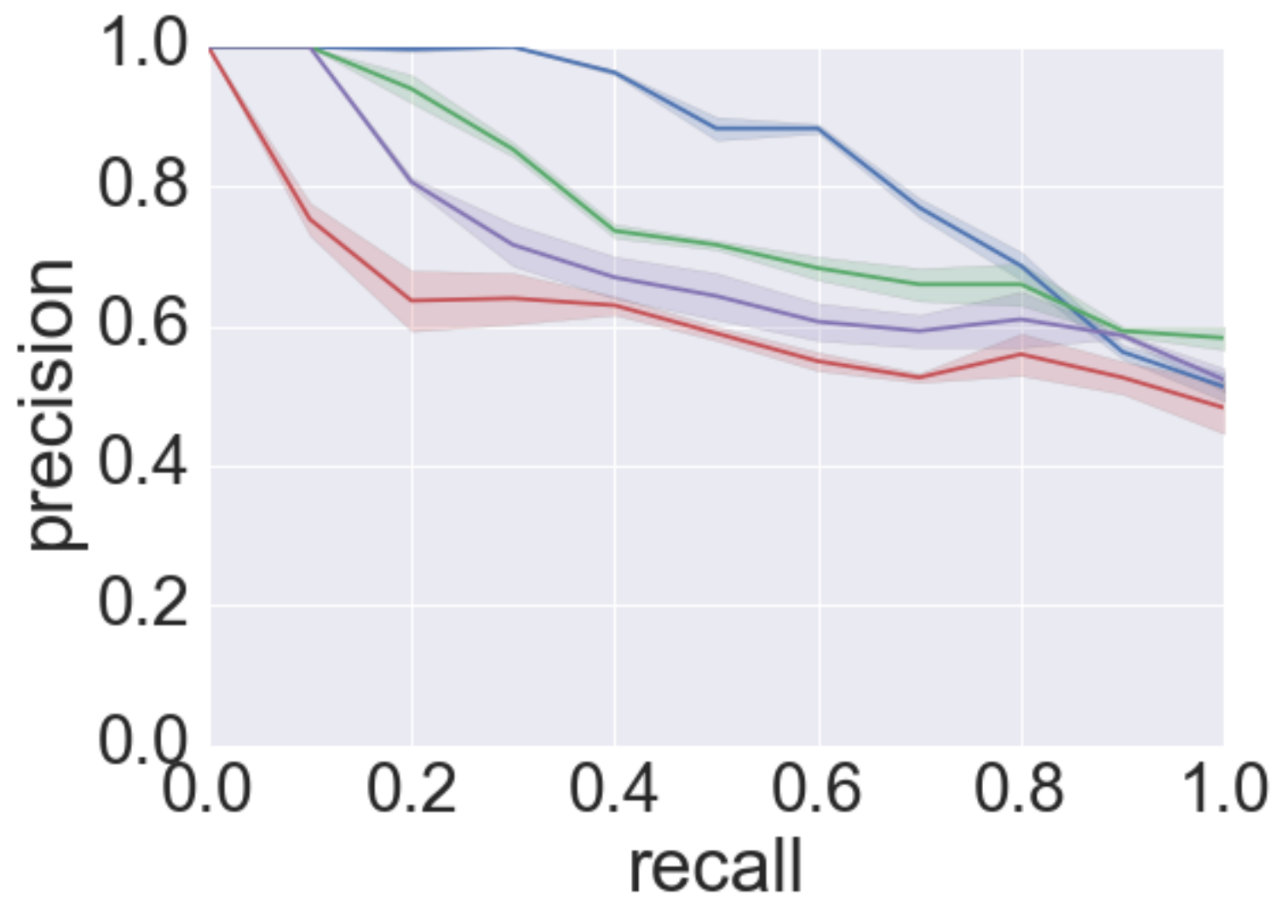}
\end{minipage}\hfill
\begin{minipage}{\figwidth} \centering
\includegraphics[width=1.0\linewidth]{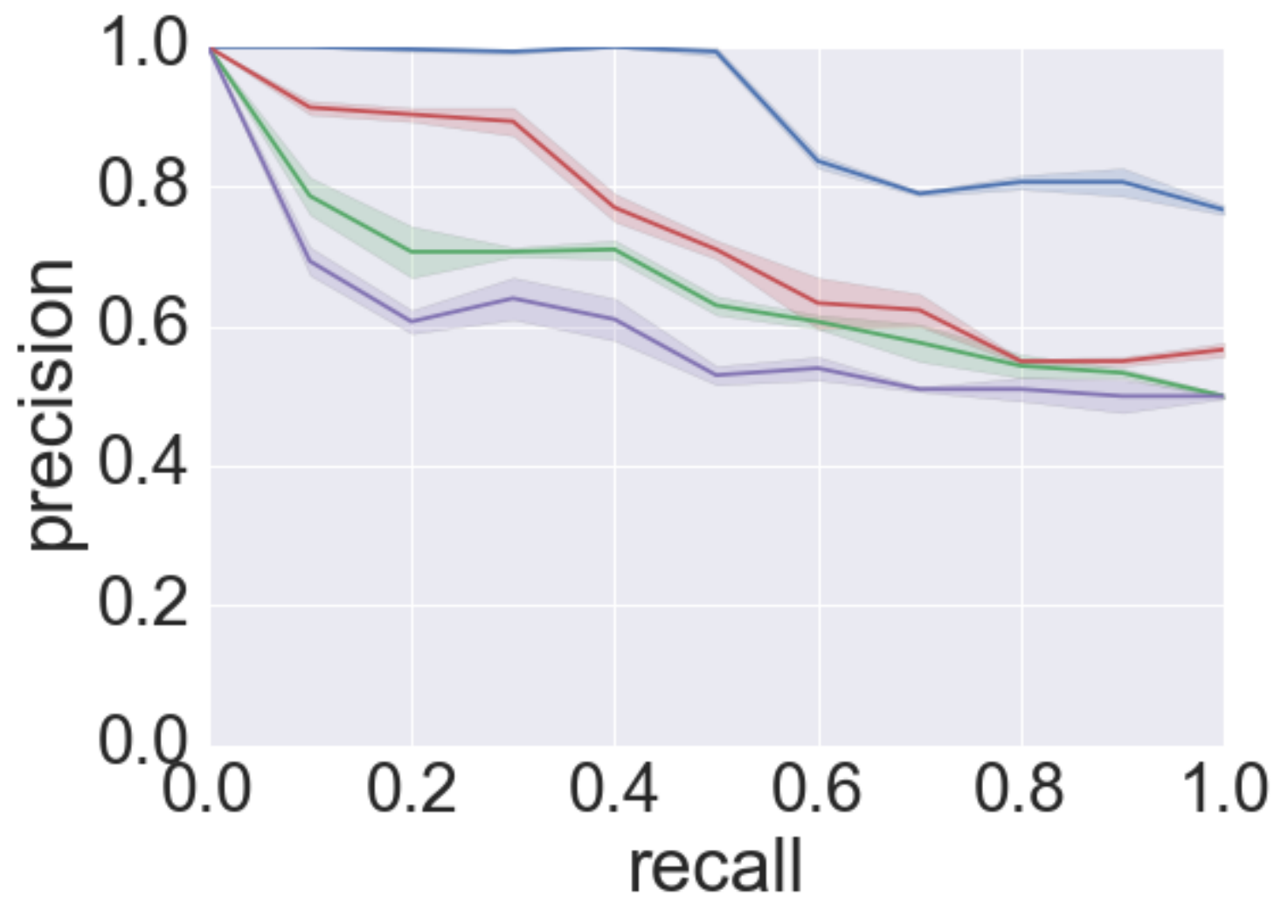} 
\end{minipage}\hfill
\begin{minipage}{\figwidth} \centering
\includegraphics[width=1.0\linewidth]{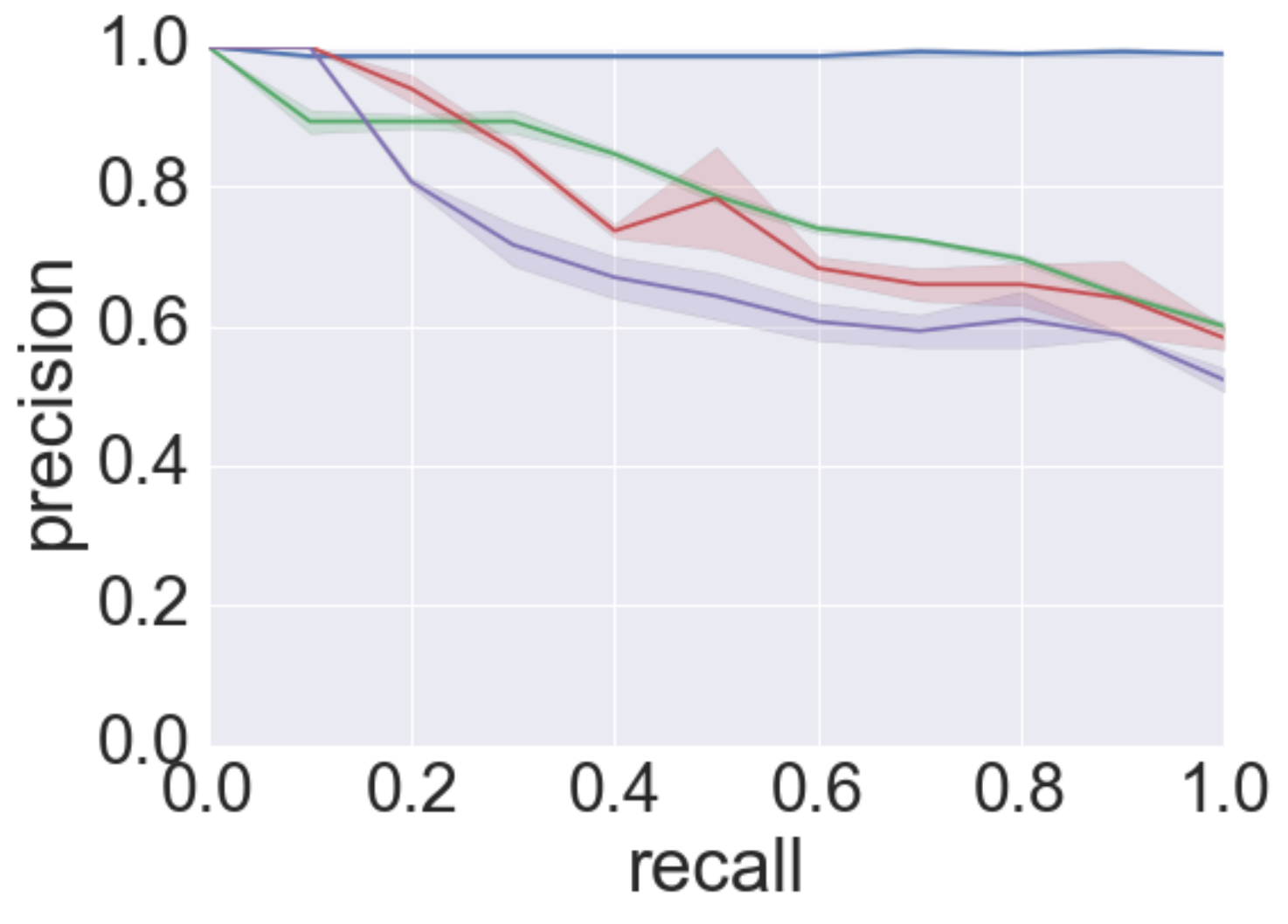} 
\end{minipage}\hfill
\begin{minipage}{\figwidth} \centering
\includegraphics[width=1.0\linewidth]{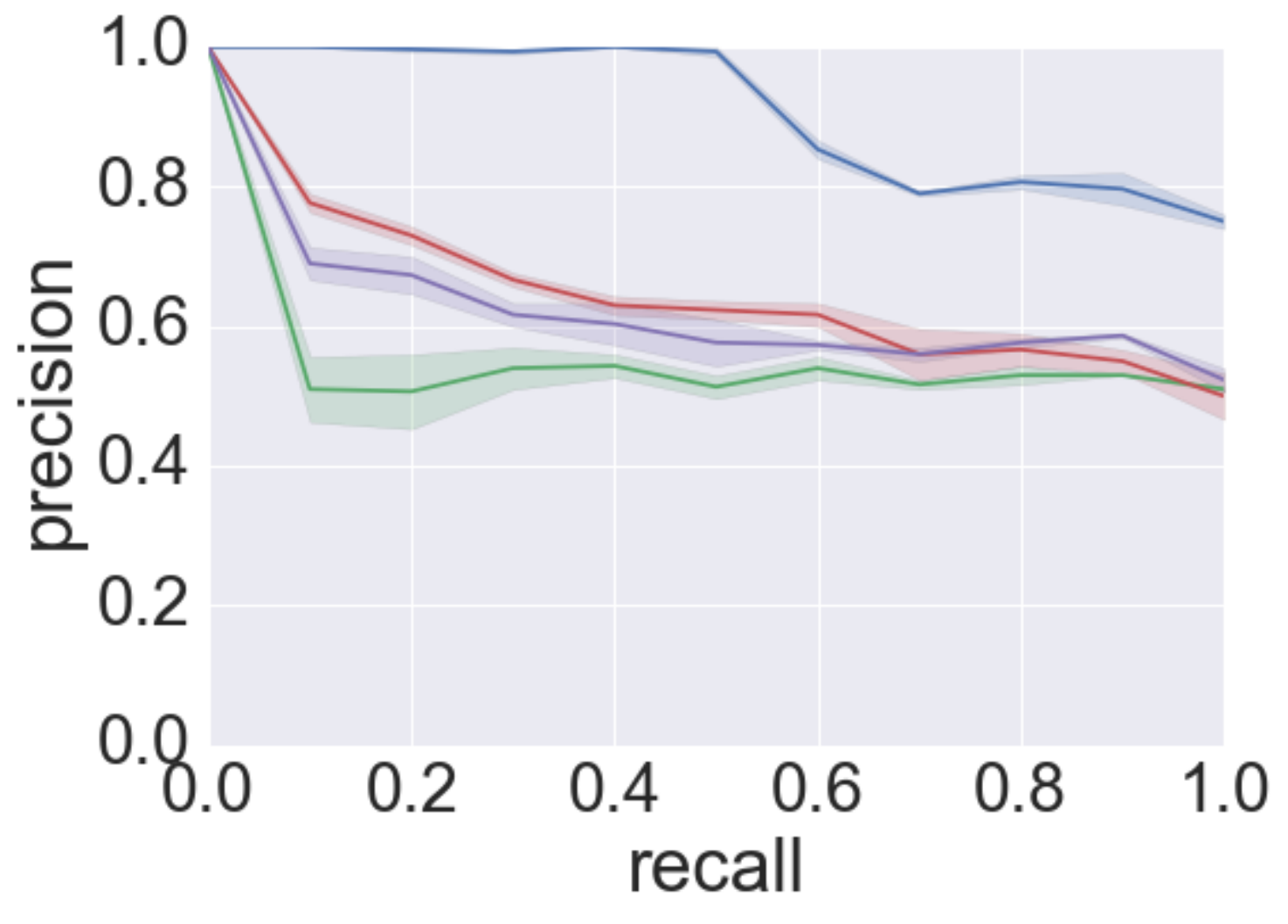} 
\end{minipage}\hfill
\begin{minipage}{\figwidth} \centering
\includegraphics[width=1.0\linewidth]{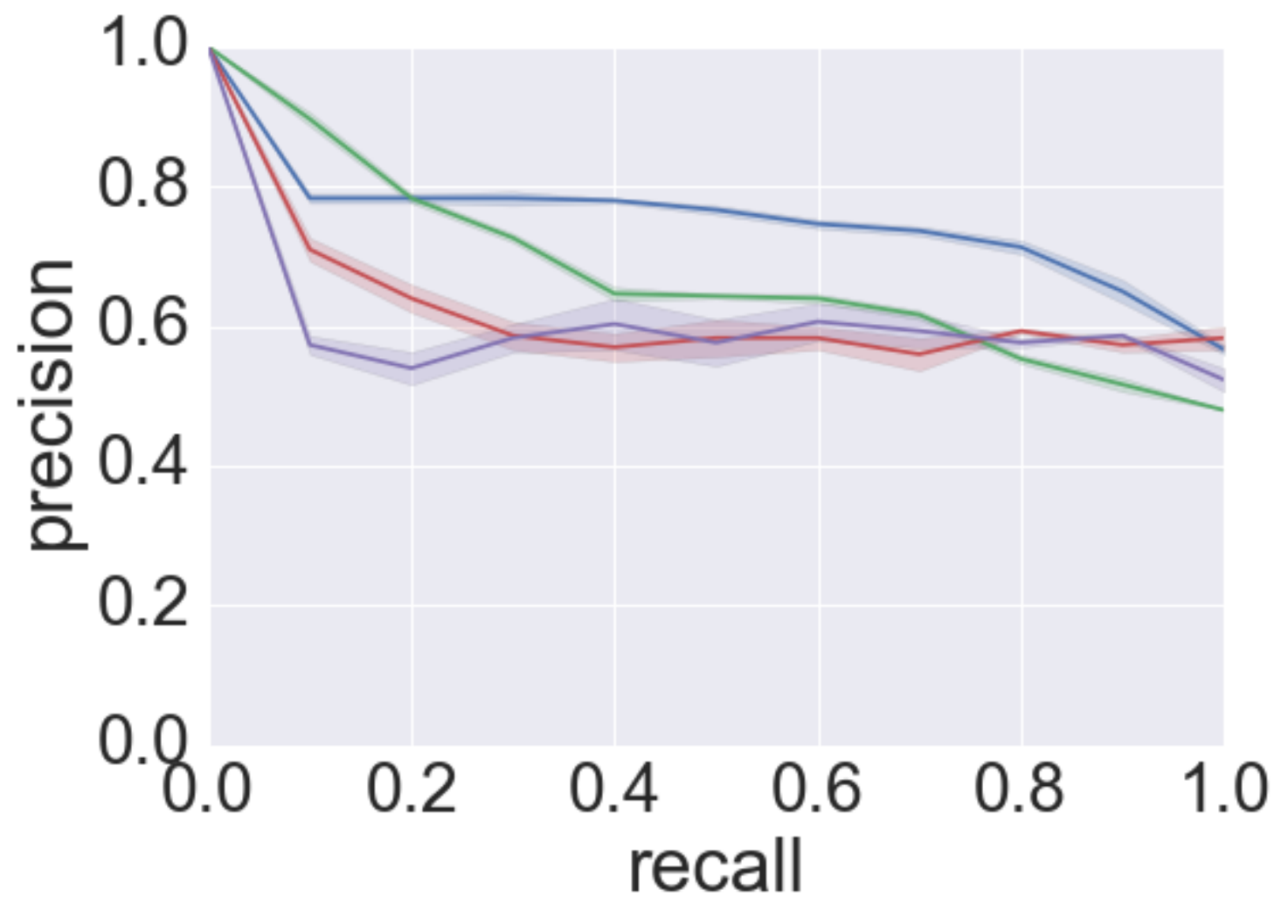}
\end{minipage}\hfill
\\
\hspace{3mm} \hfill
\begin{minipage}{\figwidth} \centering
\textsc{Pong}
\end{minipage}\hfill
\begin{minipage}{\figwidth} \centering
\textsc{Seaquest}
\end{minipage}\hfill
\begin{minipage}{\figwidth} \centering
\textsc{Freeway}
\end{minipage}\hfill
\begin{minipage}{\figwidth} \centering
\textsc{ChopperCommand}
\end{minipage}\hfill
\begin{minipage}{\figwidth} \centering
\textsc{MsPacman}
\end{minipage}\hfill

\caption{
\tb{Quantitative evaluation on detecting adversarial attacks.} Precision-recall curves in five Atari games. 
First row: Fast Gradient Sign Method (FGSM)~\cite{goodfellow:explaining}, Second row: Basic Iterative Method (BIM)~\cite{kurakin:adversarial}, and Carlini~\etal~\cite{carlini-wagner:towards}
Note that in computing the precision and recall we consider only \emph{successful} adversarial attacks (\ie attacks that do change the action decision). Including unsuccessful attack attempts may produce unreasonably high performance and thus does not reveal the actual accuracy of the detector. 
Our approach performs favorably against over several baseline detectors that uses single input.
Defense methods: \crule[blue]{0.3cm}{0.3cm} Ours, 
\crule[green]{0.3cm}{0.3cm} Feature Squeezer~\cite{xu:feature},
\crule[red!50!blue]{0.3cm}{0.3cm} AutoEncoder~\cite{dongyu:magnet}, and
\crule[red]{0.3cm}{0.3cm} Dropout~\cite{feinman:detecting}.
}
\label{fig:pr}
\end{figure*}

\setlength{\figwidth}{0.33 \textwidth}
\begin{figure*}[t!] \vspace{-2mm}
\centering
\small
\begin{minipage}{\figwidth} \centering
\includegraphics[width=1.0\linewidth]{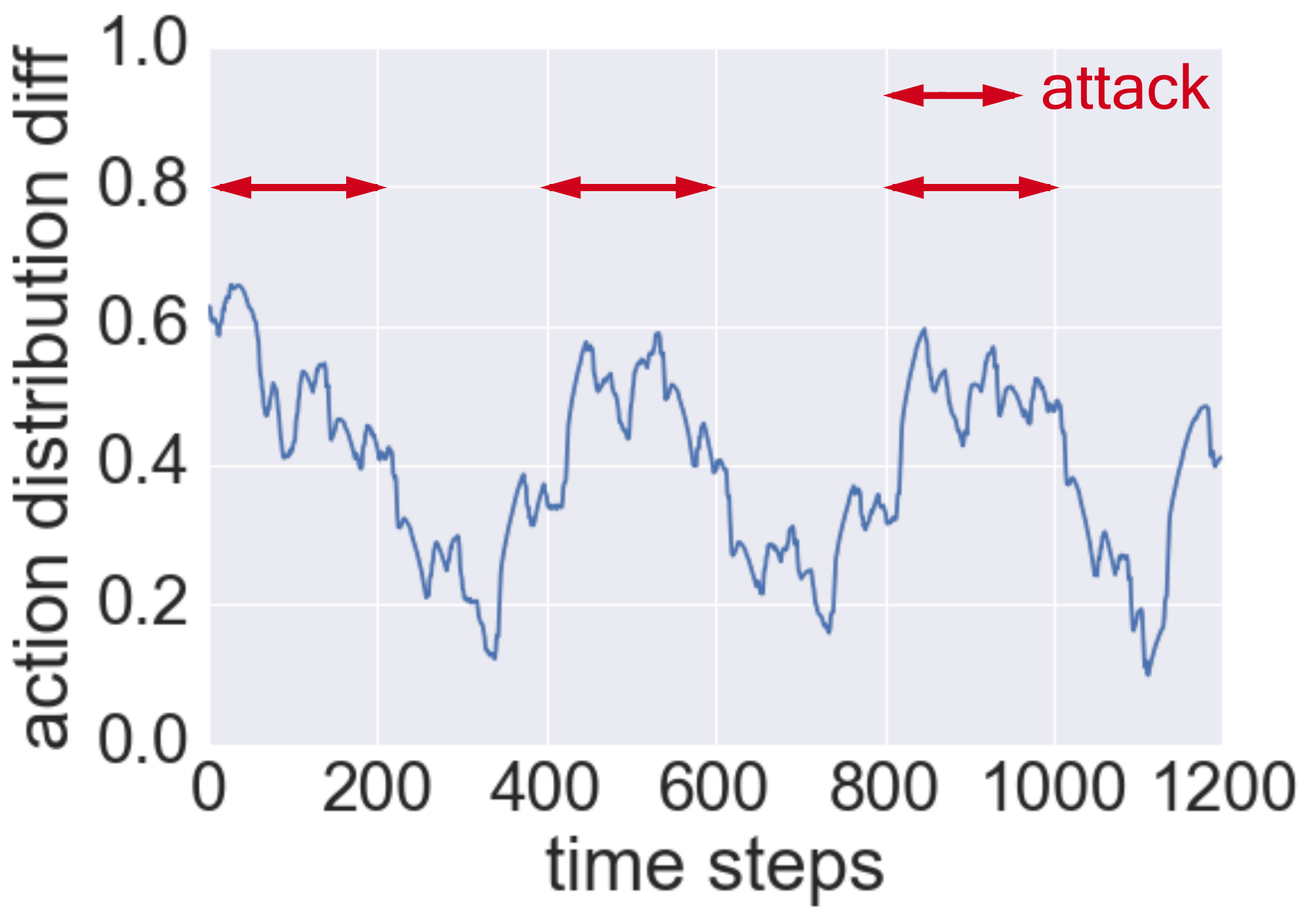} \\
\textsc{Pong}
\end{minipage}\hfill
\begin{minipage}{\figwidth} \centering
\includegraphics[width=1.0\linewidth]{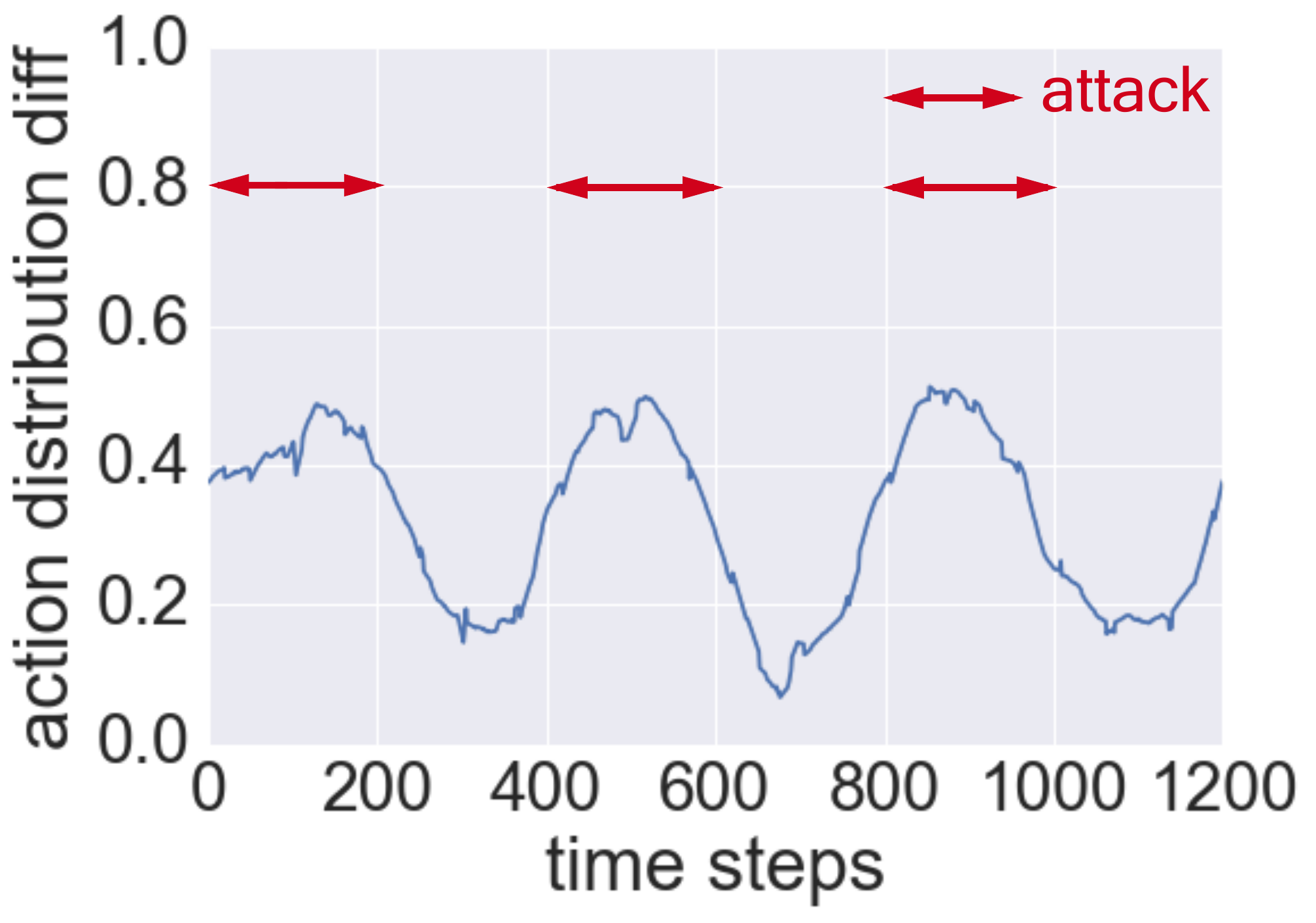} \\
\textsc{Freeway}
\end{minipage}\hfill
\begin{minipage}{\figwidth} \centering
\includegraphics[width=1.0\linewidth]{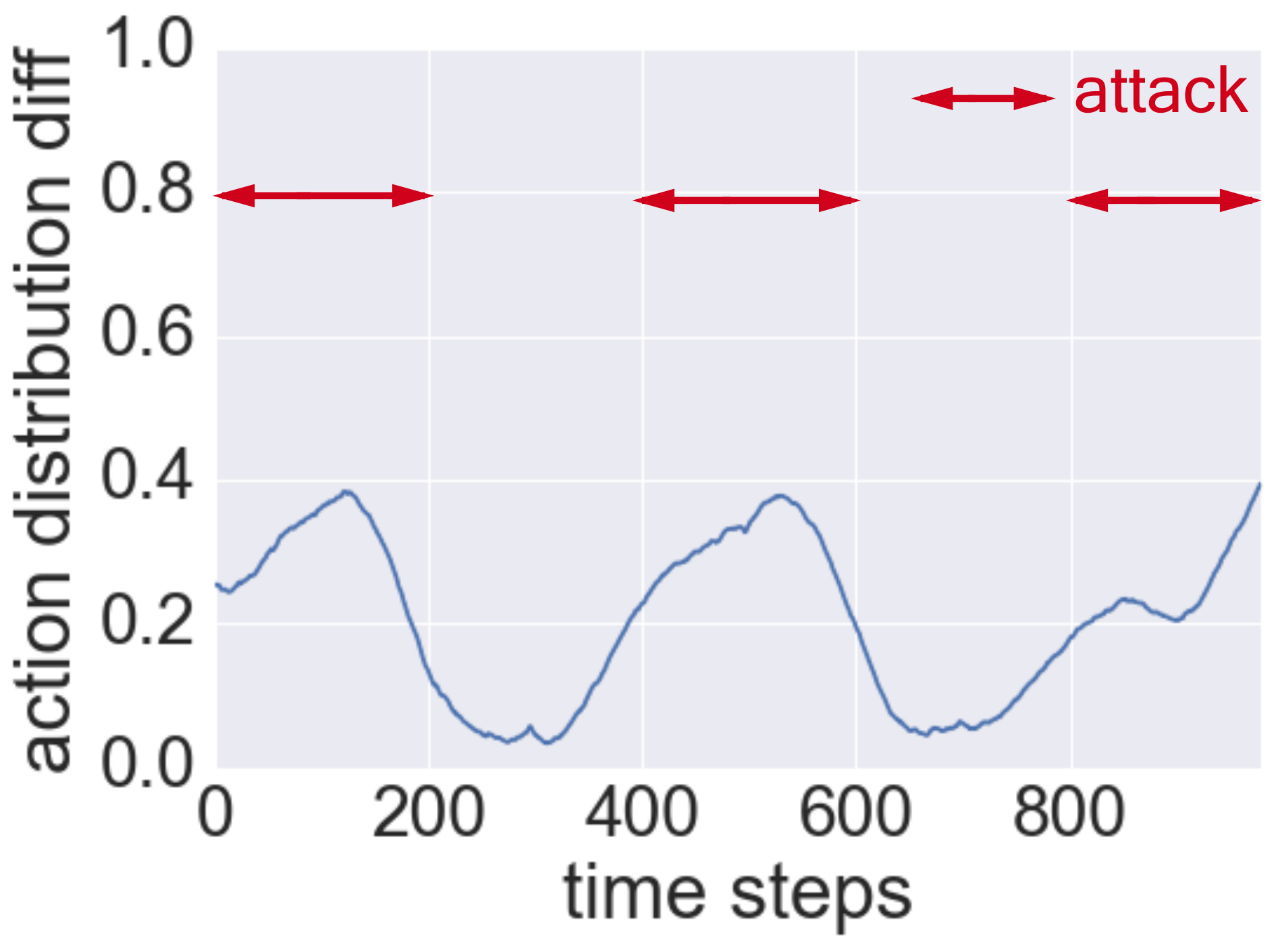} \\
\textsc{MsPacman}
\end{minipage}
\vspace{\figmargin}
\caption{
\tb{The action distribution distance with and without adversarial attacks.} For each of the three games, we visualize the action distribution distance $D(\pi_{\theta_{\pi}}(\hat{x_{t}}), \pi_{\theta_{\pi}}(x_{t}))$ across all the time steps. In this experiment, we apply adversarial attacks periodically for 200 frames as indicated by the red double arrows. The results show that distance correlate well with time windows under attack, suggesting that we can leverage it as a discriminative signal for detecting adversarial attacks.
}
\label{fig:action_distance}
\end{figure*}

\vspace{\secmargin}
\subsection{Action Suggestion} \label{sec:reward} 

\setlength{\figwidth}{0.2 \textwidth}
\begin{figure*}[t!]
\centering
\small
\begin{minipage}{\figwidth} \centering
\includegraphics[width=1.0\linewidth]{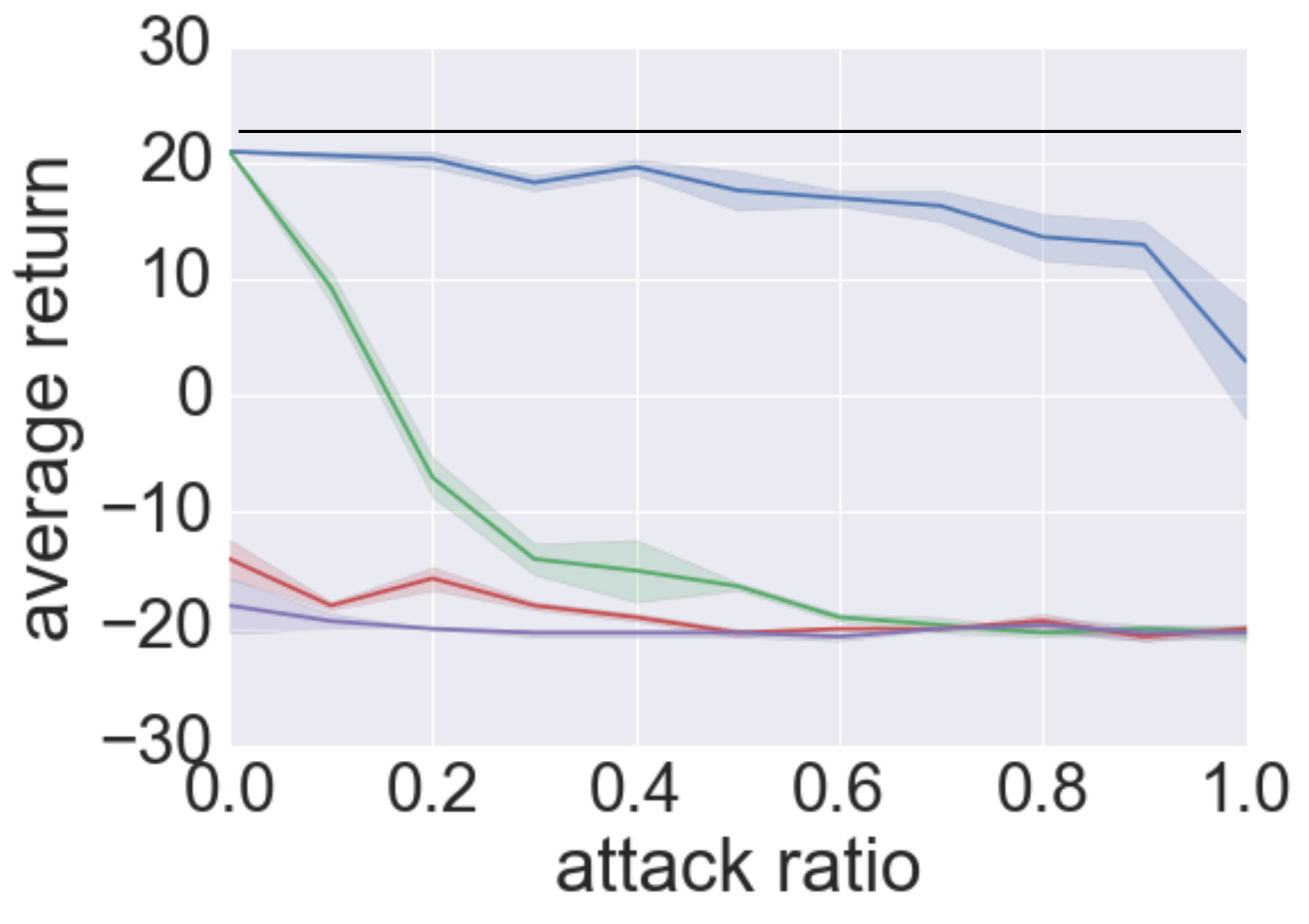} \\
\textsc{Pong}
\end{minipage}\hfill
\begin{minipage}{\figwidth} \centering
\includegraphics[width=1.0\linewidth]{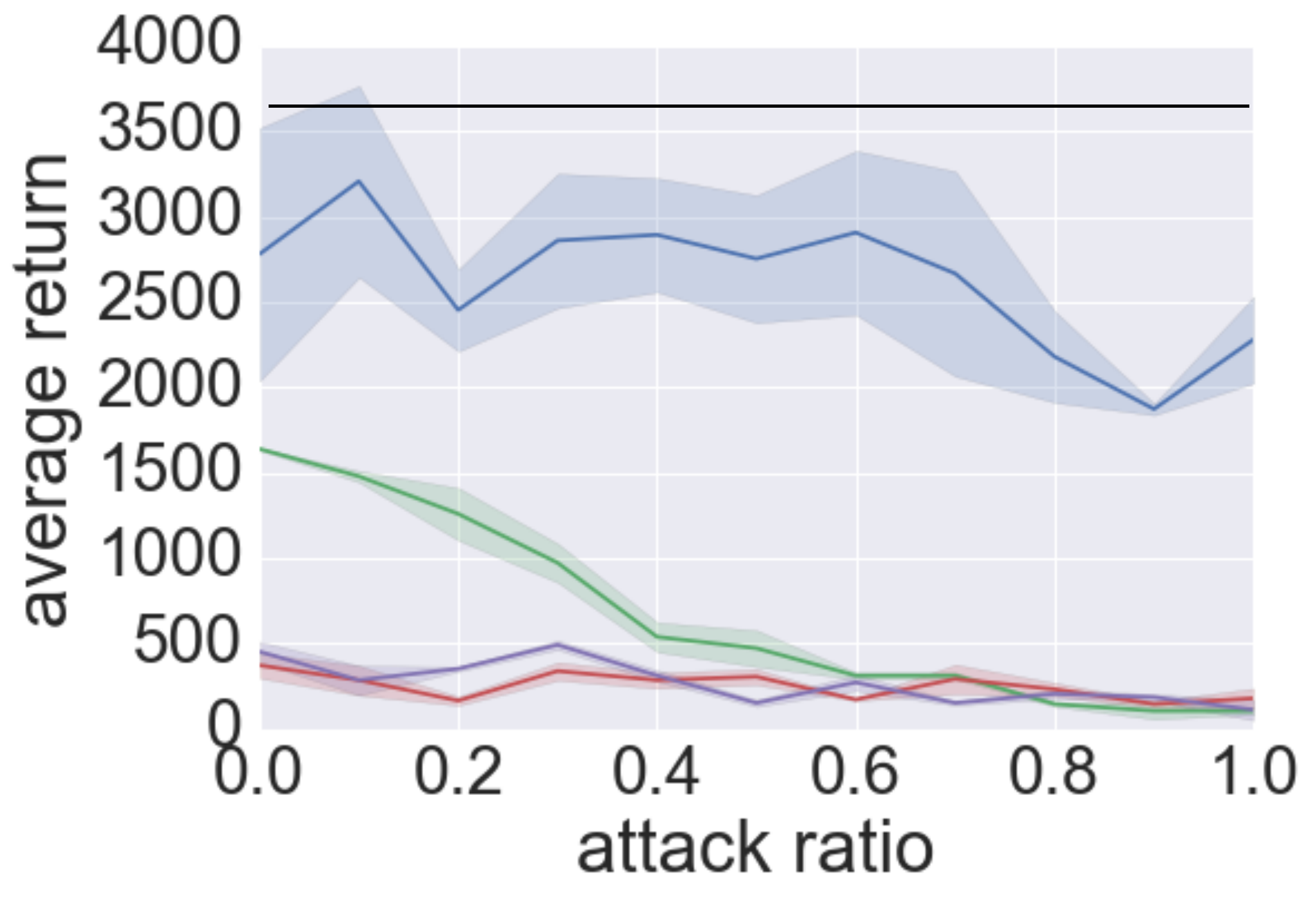} \\
\textsc{Seaquest}
\end{minipage}\hfill
\begin{minipage}{\figwidth} \centering
\includegraphics[width=1.0\linewidth]{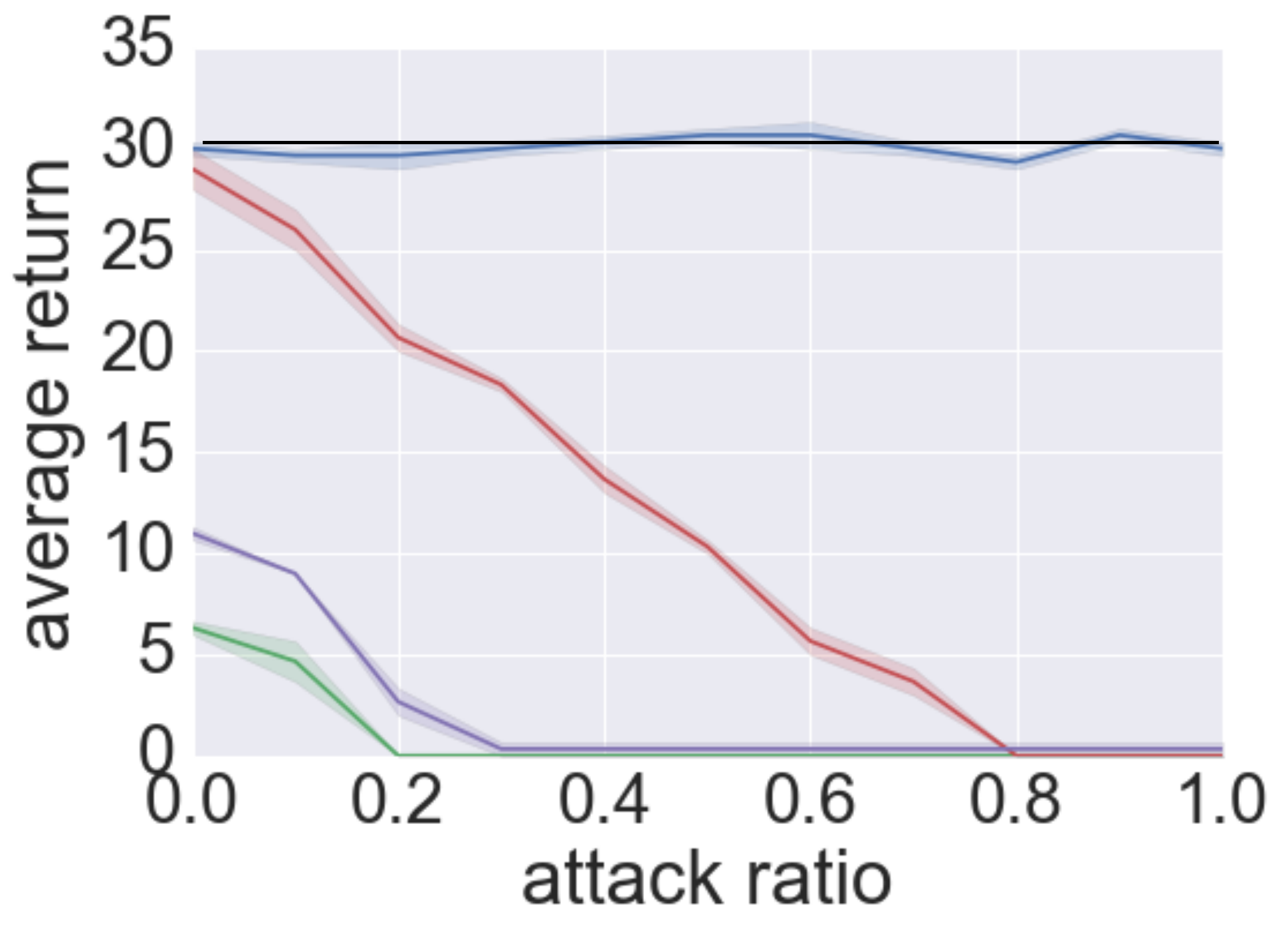} \\
\textsc{Freeway}
\end{minipage}\hfill
\begin{minipage}{\figwidth} \centering
\includegraphics[width=1.0\linewidth]{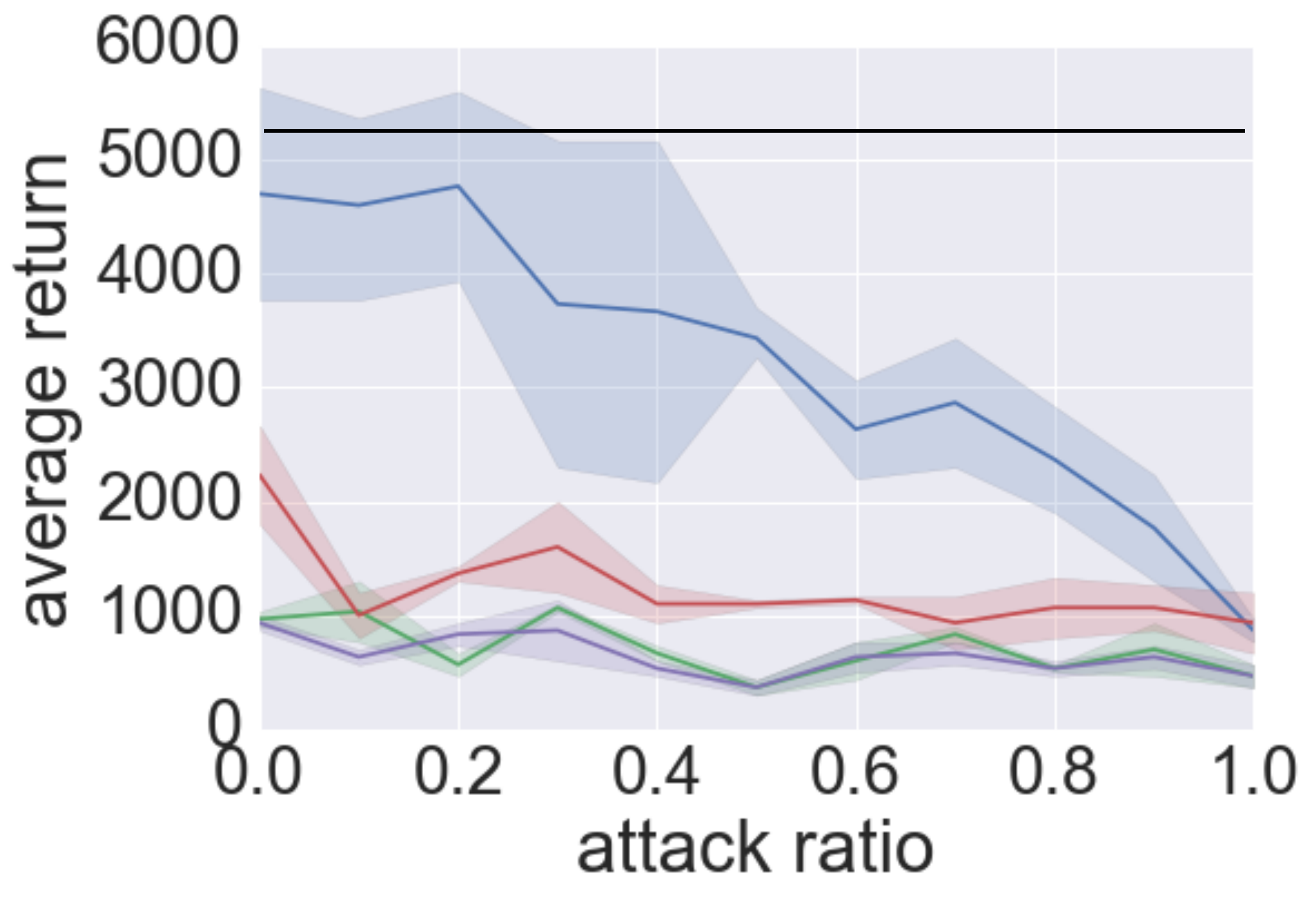} \\
\textsc{ChopperCommand}
\end{minipage}\hfill
\begin{minipage}{\figwidth} \centering
\includegraphics[width=1.0\linewidth]{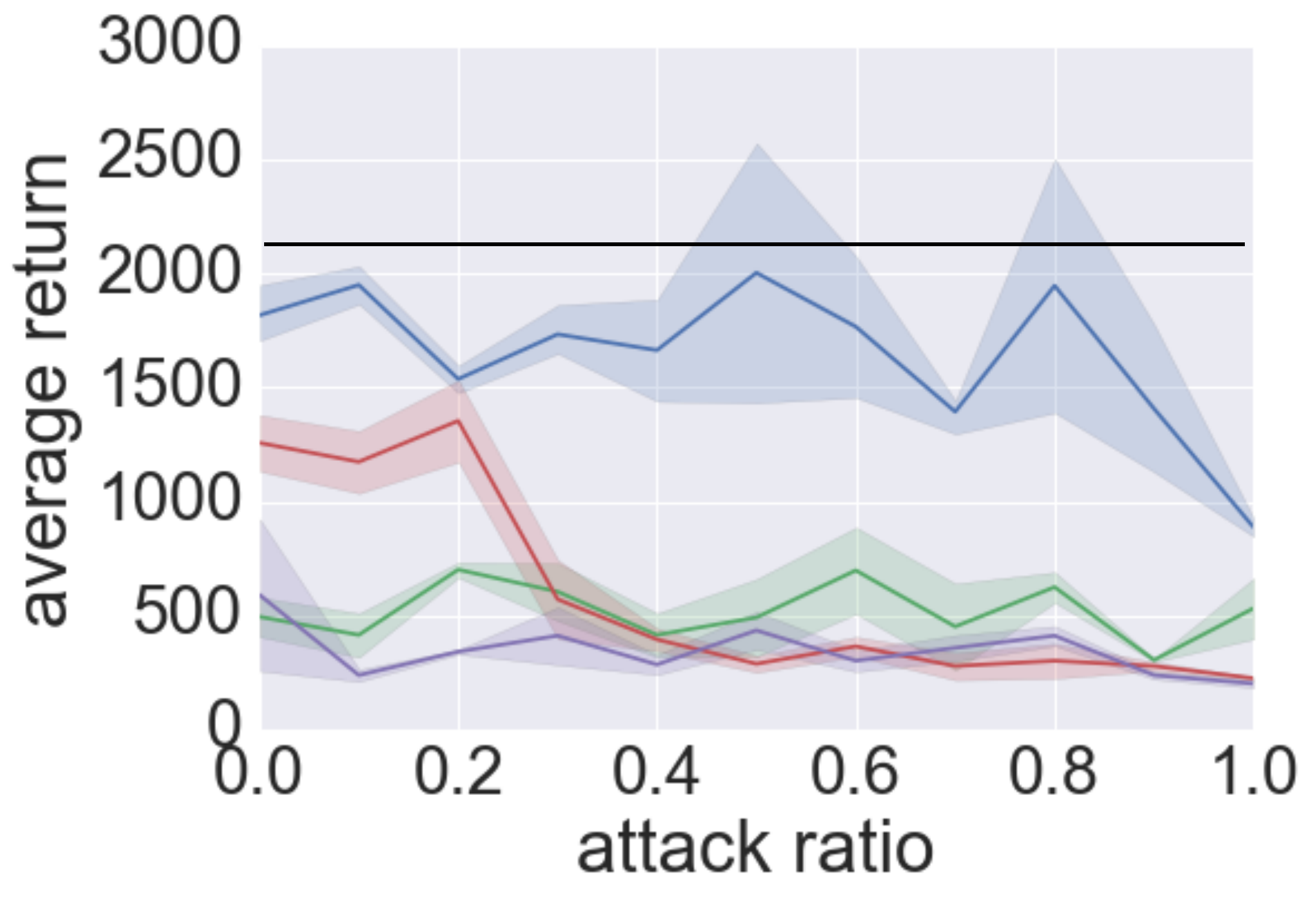} \\
\textsc{MsPacman}
\end{minipage}
\vspace{\figmargin}
\caption{\tb{Cumulated rewards under different attack ratio.} The x-axis indicates the ``attack ratio" --- the percentage of the time steps where we apply adversarial attacks. For example, when the attack ratio equals one, it indicates that the adversary applies attacks for every time step. By exploiting the visual foresight module, we can retain the agent's performance even when a large portion of the time steps are attacked by an adversary. Other alternatives such as taking random actions when the agent is under attack are not able to retain competitive performance. Defense methods: \crule[blue]{0.3cm}{0.3cm} Ours \crule[red]{0.3cm}{0.3cm} Ours-random \crule[green]{0.3cm}{0.3cm} Feature Squeezer~\cite{xu:feature} \crule[red!50!blue]{0.3cm}{0.3cm} Feature Squeezer-random
}
\label{fig:reward}
\end{figure*}
In \figref{reward}, we evaluate the effectiveness of the proposed action suggestion method. 
We vary the attack ratio in an episode in the range of $[0, 1]$ and show the averaged rewards in a total five trials for each game.
The baseline detector by~\cite{xu:feature} could also provide action suggestions using the \emph{filtered} current frame as input to the policy.
We also evaluate ``random actions" as another alternative baseline for action suggestions. 
When the agent detects an attack at a time step, it selects a random action to perform.
We also show the rewards obtained by the same agent when there are no adversarial attacks (shown as the solid black lines). 
Our experiment results demonstrate that the agent is capable of retaining good performance through the action suggestions even when a large portion of the time steps were attacked.

\vspace{\secmargin}
\subsection{The Effect of Frame Prediction Model}
\label{sec:effect_frame_prediction}

\setlength{\figwidth}{0.2 \textwidth}
\begin{figure}[t!] 
\vspace{\figmargin}
\centering
\small
\includegraphics[width=0.5\linewidth]{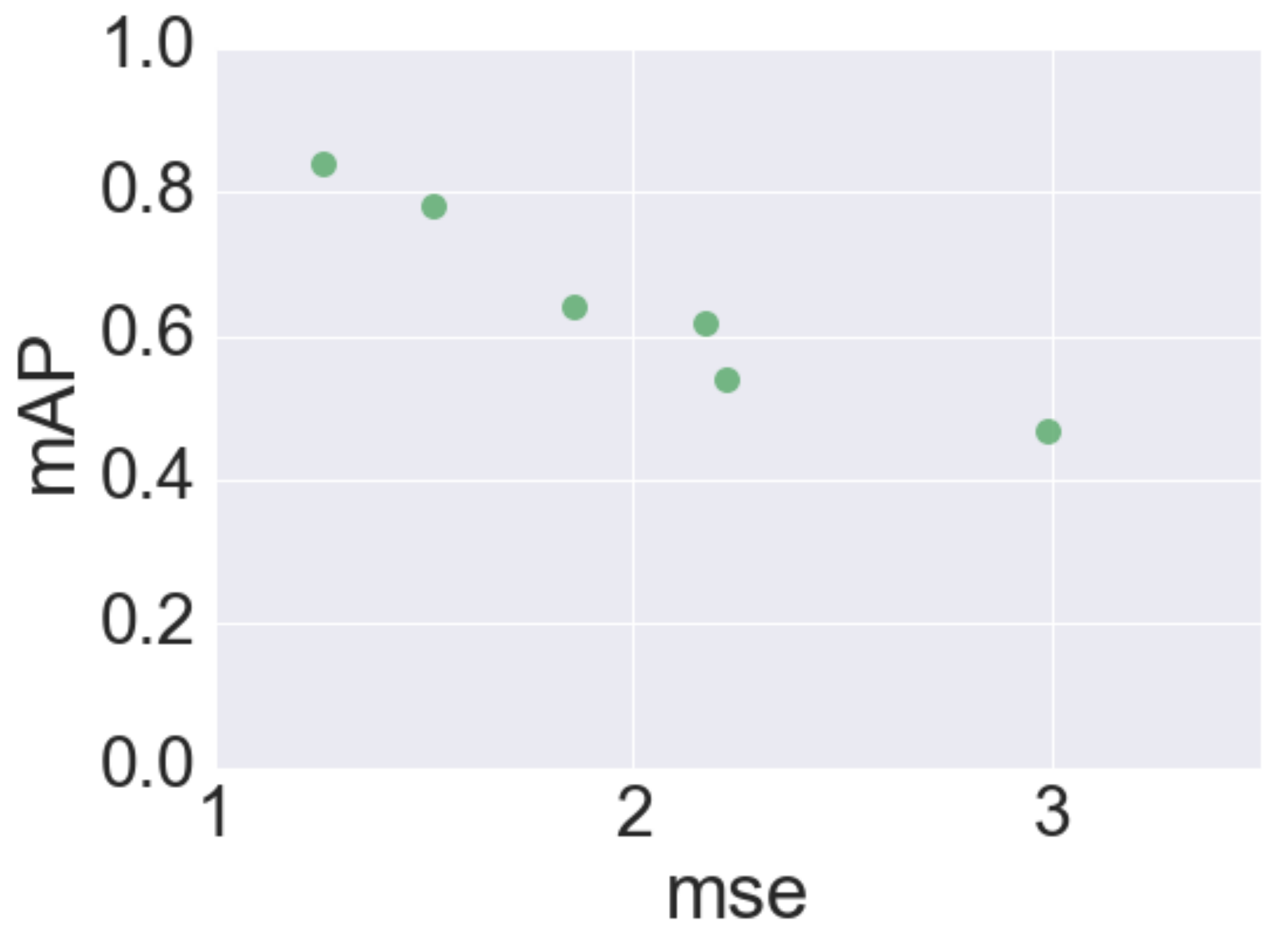}\\
\caption{
\tb{Detector performance vs. the accuracy of the visual foresight module.} Using a more accurate frame prediction model yields improved performance in detecting adversarial examples.
}
\label{fig:quality}
\end{figure}


\ignorethis{
\begin{figure*}[t!]\vspace{-2mm}
\centering
\begin{subfigure}{0.5\textwidth}
  \centering
  \includegraphics[width=1.0\linewidth]{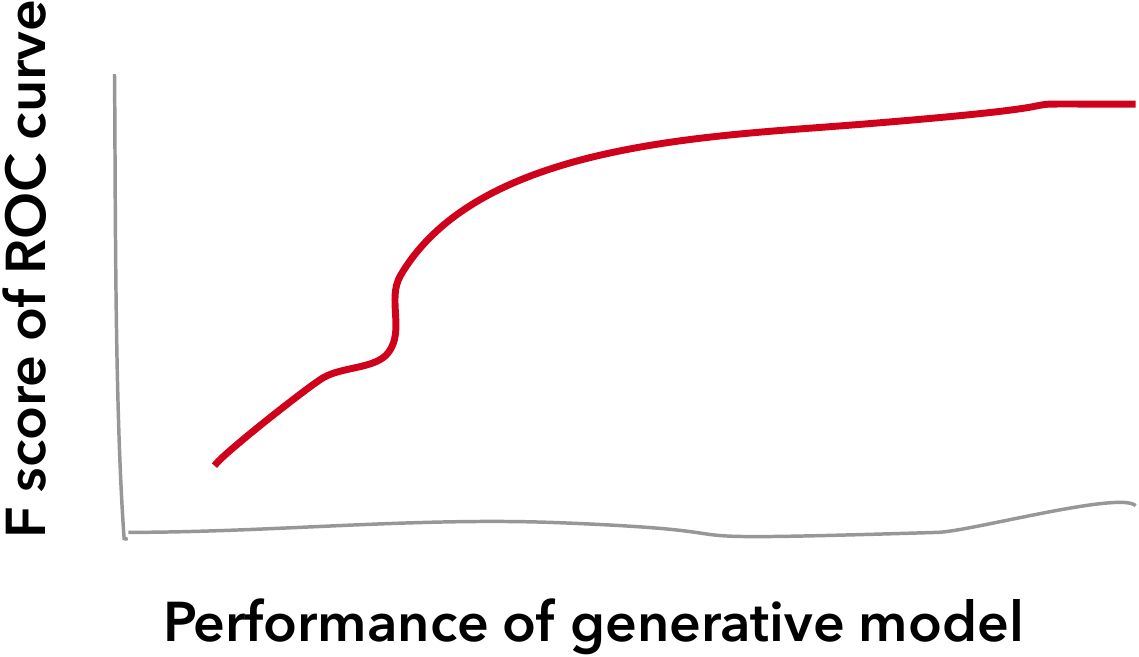}
  \caption{Pong}
  \label{fig:sfig1}
\end{subfigure}

\ignore{
\begin{subfigure}{.2\textwidth}
  \centering
  \includegraphics[width=0.8\linewidth]{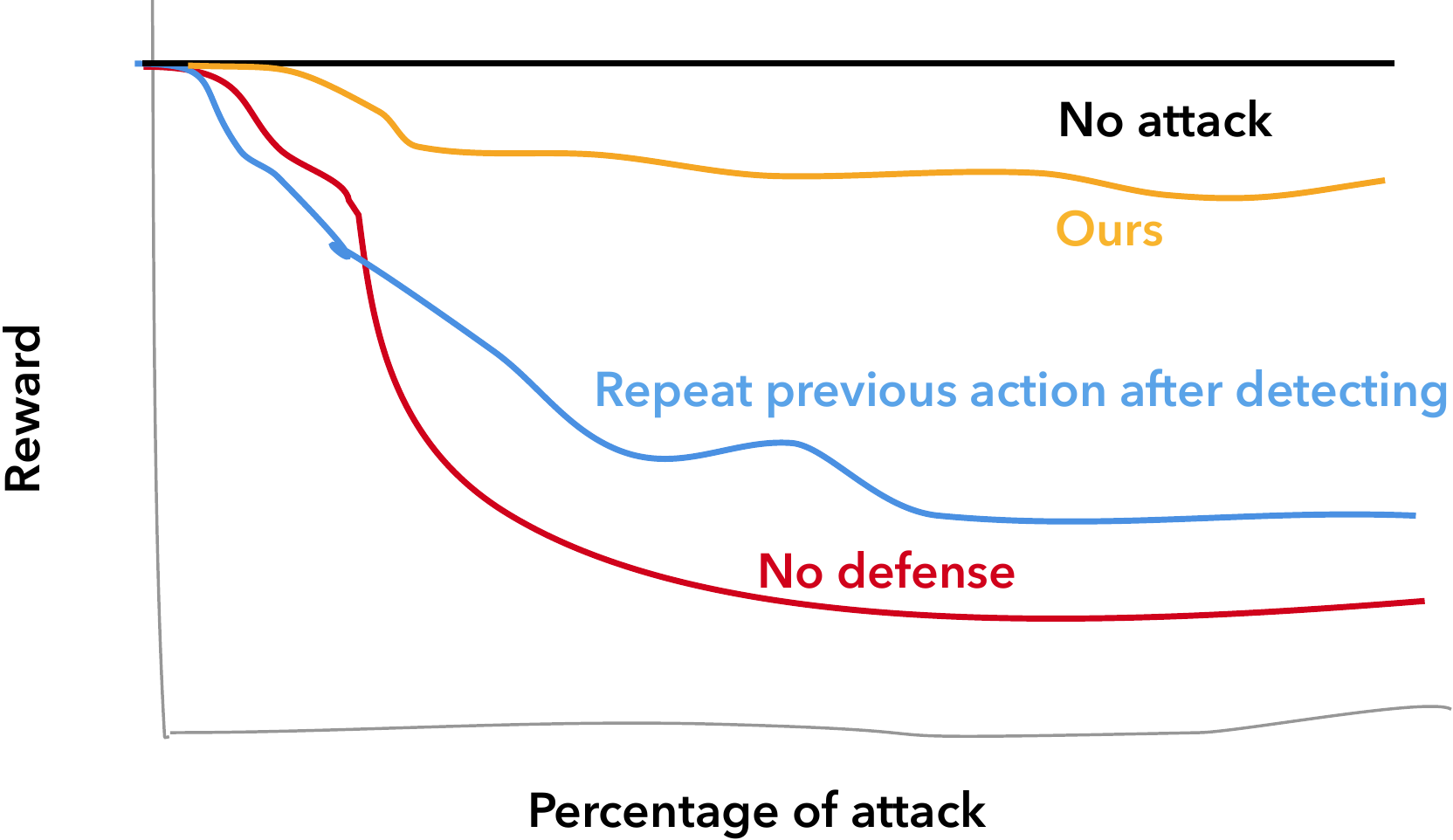}
  \caption{Seaquest}
  \label{fig:sfig2}
\end{subfigure}%
\begin{subfigure}{.2\textwidth}
  \centering
  \includegraphics[width=0.8\linewidth]{./figures/reward.pdf}
  \caption{MsPacman}
  \label{fig:sfig3}
\end{subfigure}%
\begin{subfigure}{.2\textwidth}
  \centering
  \includegraphics[width=0.8\linewidth]{./figures/reward.pdf}
  \caption{ChopperCommand}
  \label{fig:sfig4}
\end{subfigure}%
\begin{subfigure}{.2\textwidth}
  \centering
  \includegraphics[width=0.8\linewidth]{./figures/reward.pdf}
  \caption{Freeway}
  \label{fig:sfig5}
\end{subfigure}\vspace{-2mm}
}
\vspace{\figmargin}
\caption{F score of ROC curve (y-axis) v.s. $X$ reconstruction error of generative model (x-axis) in 5 games.}
\label{fig:quality}
\end{figure*}
}

We investigate the effect of the action-conditioned frame prediction model.
During the training process, we take model snapshots at 40K, 80K, 120K, 160K, 320K, and 640K iterations to obtain frame prediction models with varying degrees of prediction accuracies.
We take the game \textsc{Pong} as an example. 
We run the policy equipped with different frame prediction models. 
In \figref{quality}, we show the scatter plot showing the relationship between 
1) x-axis: the quality of the frame prediction model (measured in terms of mean squared error) and 
2) y-axis: the effectiveness of the detector (measured in term of mean Average Precision (mAP)).
The results show that the accuracy of the frame prediction model plays a critical role for detecting adversarial examples.

\vspace{\secmargin}
\section{Discussions}

In this section, we discuss the main limitations of the proposed defense mechanism and describe potential future directions for addressing these issues.

\vspace{\paramargin}
\paragraph{Defense against an adaptive adversary. } 
In this paper, we only consider the case of \emph{static} adversarial attack.
Defending against \emph{adaptive} adversary (who knows about the defense being used) remains an open problem in the field~\cite{carlini-wagner:adversarial}.
%
We believe that leveraging the temporal coherence and redundancy may help construct an effective defense against strong, adaptive adversaries.
Here we explain such a potential extension. 
Our detector leverage previous observations for detecting adversarial examples.
As the agent store these previous observations in its memory, 
the adversary \emph{cannot} change the previous observations perceived by the agent at the current time step $t$.
Such a design poses an additional challenge for an adaptive adversary to attack our defense. 
As our frame prediction model takes $m$ previous observation as inputs, an adaptive adversary can potentially exploit two strategies. 
First, the adaptive adversary can start applying the attack at time step $t-m$ in order to fool the frame prediction model (in addition to fooling the agent).
In such cases, however, our defense will force the adversary to apply the attack in more time steps than needed and therefore render the adversary easier to be spotted.
Second, an adaptive adversary may augment its adversarial example generation process with a frame prediction module and craft adversarial examples that cause our frame prediction model to predict an adversarial example for the agent.
To the best of our knowledge, no existing frame prediction models can predict adversarial examples for another network (\eg the DQN~\cite{mnih:human} in this paper). 

\vspace{\paramargin}
\paragraph{Practical considerations for applying adversarial attack in the physical world. }
In the real world, robotics often operate in real time. 
Therefore, attacking a robotic system using adversarial examples generated by iterative methods may become impractical as iterative methods are generally computational expensive.
Existing works for crafting adversarial examples for image classification tasks often do not take the computational cost into consideration when designing an attack algorithm.
In the context of sequential decision-making task in the physical world, we believe that the efforts for increasing the time required by an adversary to craft an effective adversarial example are promising directions for securing these systems.

\vspace{\paramargin}
\paragraph{Frame prediction in the physical world.} Our defense performance relies on the accuracy of the action-conditioned frame prediction model as shown in \figref{quality}.
%
Therefore, our method is applicable for controlled settings in the physical world (\eg in-house robotics~\cite{finn:deep}) and simulated environment (\eg games~\cite{oh:action}) where we can train an accurate action-conditioned frame prediction model.
%
Extending it to the \emph{unconstrained} setting in the physical world is not trivial because of the vast appearance variations and uncertainty of the immediate future. 
One promising extension is to simultaneously train the DNN-based policy and the frame prediction model so that the agent can accommodate the artifacts produced by the frame prediction model.

\vspace{\paramargin}
\paragraph{Integration with other techniques.}
We note that the proposed defense leverages temporal coherence for detecting adversarial examples. 
Such temporal information is orthogonal with existing methods that exploit information extracted from a  single input. 
Consequently, our approach can potentially be integrated with adversarial training~\cite{goodfellow:explaining}, defensive distillation~\cite{papernot:distillation} and other adversarial examples detection algorithms to construct a stronger defense.


\ignorethis{
\subsection{Design Implication}
One critical difference between our method and existing methods is that instead of detecting adversarial examples from one single observation, our detector also leverage previous observations. 
As the agent store those previous observations in its memory, the adversary \emph{cannot} change the previous observations at the current time step $t$.
\jiabin{What's that additional challenge?}
Such a design poses an additional challenge for the adversary. 
}
\ignorethis{
The incorporation dropout in our frame prediction model during run-time makes it more difficult to generative an effective adversarial example within a reasonably short period of time.
}
\section*{Acknowledgements}
We gratefully acknowledge the support of NVIDIA Corporation with the donation of the Titan X GPU used for this research.


\bibliographystyle{ieee}
\bibliography{Example}  

\end{document}